\pdfoutput=1
\documentclass[sigconf]{acmart}

\hypersetup{colorlinks,allcolors=black}

\copyrightyear{2020}
\acmYear{2020}
\setcopyright{acmcopyright}
\acmConference[AISec '20] {13th ACM Workshop on Artificial Intelligence and Security}{November 13, 2020}{Virtual Event, USA}
\acmBooktitle{13th ACM Workshop on Artificial Intelligence and Security (AISec '20), November 13, 2020, Virtual Event, USA}
\acmPrice{15.00}
\acmDOI{10.1145/3411508.3421380}
\acmISBN{978-1-4503-8094-2/20/11}

\setcitestyle{numbers}
\usepackage[linesnumbered,ruled,vlined]{algorithm2e}
\usepackage{graphicx}
\usepackage{amsmath}
\usepackage{multirow}
\usepackage{caption}
\usepackage[normalem]{ulem}
\useunder{\uline}{\ul}{}
\usepackage{subcaption}
\AtBeginDocument{%
\providecommand\BibTeX{{%
\normalfont B\kern-0.5em{\scshape i\kern-0.25em b}\kern-0.8em\TeX}}}

\newcommand{\itrans}{irreversible transformation}
\newcommand{\rtrans}{reversible transformation}

\usepackage[british]{babel}
\renewcommand\footnotetextcopyrightpermission[1]{} 
\begin{document}
\fancyhead{}

\title{Where Does the Robustness Come from? A Study of the Transformation-based Ensemble Defence}


\author{Chang Liao}
\authornote{This work was done during his internship at Huawei International, Singapore.}
\email{cliao002@e.ntu.edu.sg}
\affiliation{%
 \institution{Nanyang Technological University}
 \city{Singapore}
}

\author{Yao Cheng}
\authornote{Corresponding author.}
\email{chengyao101@huawei.com}
\affiliation{%
 \institution{Huawei International}
 \city{Singapore}
}

\author{Chengfang Fang}
\email{fang.chengfang@huawei.com}
\affiliation{%
 \institution{Huawei International}
 \city{Singapore}
}

\author{Jie Shi}
\email{shi.jie1@huawei.com}
\affiliation{%
 \institution{Huawei International}
 \city{Singapore}
}




\begin{abstract} 
This paper aims to provide a thorough study on the effectiveness of the transformation-based ensemble defence for image classification and its reasons. It has been empirically shown that they can enhance the robustness against evasion attacks, while there is little analysis on the reasons. In particular, it is not clear whether the robustness improvement is a result of transformation or ensemble. In this paper, we design two adaptive attacks to better evaluate the transformation-based ensemble defence. We conduct experiments to show that 1) the transferability of adversarial examples exists among the models trained on data records after different reversible transformations; 2) the robustness gained through transformation-based ensemble is limited; 3) this limited robustness is mainly from the irreversible transformations rather than the ensemble of a number of models; and 4) blindly increasing the number of sub-models in a transformation-based ensemble does not bring extra robustness gain. 
\end{abstract}




\settopmatter{printfolios=false}

\maketitle

\SetKwInput{KwInput}{Input}                
\SetKwInput{KwOutput}{Output}              

\section{Introduction}\label{sec:introduction}


Deep Neural Networks (DNNs) are widely deployed in various domains.
However, DNNs are shown to be intrinsically vulnerable to Adversarial Examples (AEs) which are the input samples maliciously crafted to mislead the victim DNN to an erroneous output. 
This is also known as the evasion attack. 
There have been various algorithms proposed to generate AEs~\cite{szegedy2013intriguing}\cite{fgsm}\cite{moosavi2016deepfool}\cite{carlini2017towards}.
In order to maintain stealthy in the evasion attack, these algorithms usually generate AEs bounded by certain distance measurements like $L_1, L_2,$ and $L_{\infty}$ norms to guarantee the generated AEs are similar to the original input and imperceptible to human eyes~\cite{carlini2017towards}.

It is important to defend against AEs so that we can trust DNNs in critical applications, such as autonomous driving~\cite{tencent2019tesla}.
Since the discovery of the evasion attack in 2013~\cite{szegedy2013intriguing}, continuous efforts have been put in defending DNNs against AEs~\cite{papernot2016distillation}\cite{xu2017feature}.
However, the adaptive attack~\cite{carlini2017towards}\cite{Carlini}, which is specifically designed to target a given defence, leads the DNN evasion attack and defence into an endless arms race. 

Among various defence mechanisms, ensemble is one of the major defence directions that attracts a wide range of attention. 
Ensemble-based defence approaches aim at gaining adversarial robustness by incorporating an ensemble of sub-models. 
Abbasi and Gagn{\'e}~\cite{abbasi2017robustness} are the first to defend evasion attacks using an ensemble of specialists. 
Though their defence soon was broken by He et al.~\cite{he2017adversarial}, many follow up research works continue to improve the ensemble technique~\cite{rse}\cite{adp}\cite{ECNN}\cite{ECOC}\cite{meng2020ensembles}.

The key point in an ensemble defence is the selection of the ensemble sub-models. 
The majority of the ensembles encourage their sub-models (aka, ensemble members) to be as diverse as possible.
This can be achieved by promoting the diversity among sub-models~\cite{adp}, using mixed precision of weights and activation functions~\cite{sen2020empir}, training an ensemble of binary-classifiers with sufficient diversity and redundancy~\cite{ECOC}, etc.
It is expected that the AEs can only fool a minority of the diversified sub-models.
In this manner, when the ensemble aggregates the final decision using various ensemble strategies, such as majority voting, it is more possible for it to provide a correct output than that from just one single model. 

Recently, there is a novel way of constructing an ensemble based on image transformations being proposed to enhance model robustness~\cite{meng2020ensembles}.
Instead of explicitly promoting the diversity among sub-models, this approach combines a pool of candidate sub-models associated with a large variety of transformations.
It is shown to be robust against AEs generated by various attack algorithms even when the attack has full knowledge of the ensemble including the sub-models and the ensemble strategy.
Since many ensemble defence methods have been successfully attacked soon after its proposal, it attracts our great interest to analyse \textit{where exactly the robustness comes from in the transformation-based ensembles}. 

Bearing this question in mind, in this paper, we carry out empirical experiments to analyse the robustness of the transformation-based ensemble defence. 
We notice that there are two types of transformations~\cite{meng2020ensembles}, i.e., \rtrans \  and \itrans.
We design separate experiments to evaluate the ensemble defences based on the two types of transformations, respectively.

For the ensemble based on \rtrans s, i.e., the ensemble whose sub-models are trained on the data being pre-processed by \rtrans s, two adaptive attacks are designed to better understand its robustness, i.e., Transferability-based Adaptive Attack (TAA) and Perturbation Aggregation Attack (PAA).
TAA is inspired by Liebig's law of the minimum in agricultural science that the growth is dictated not by total resources available, but by the scarcest resources~\cite{Ebelhar2008}.
Based on the transferability analysis, TAA enables us to identify the ``scarcest'' sub-model, attacking which may compromise the overall prediction of the ensemble.
We also call it ``weakest'' sub-model from the attacker's point of view.
Differently, PAA is to consider as much as possible the characteristics of the sub-models instead of just one ``weakest'' sub-model.
PAA calculates the adversarial perturbation for each sub-model and combines the perturbations using various aggregation strategies to generate a practical AE bounded by a designated dissimilarity score.

For ensemble based on \itrans s, i.e., the ensemble whose sub-models are trained on the data after \itrans s, we evaluate its robustness using transferability-based attacks~\cite{meng2020ensembles}.
In such irreversible ensembles, it is difficult to project the perturbation generated on sub-models back to the original input due to the existence of the \itrans.
Hence PAA, which aggregates individual perturbations, is not applicable to irreversible ensembles.
Moreover, in order to identify the robustness source, we control the proportion of sub-models based on \rtrans \ and \itrans \ in a hybrid ensemble to measure the contribution from the two types of transformations.
We randomly draw a controlled number of the irreversible sub-models, add them to the ensemble of reversible sub-models that has been evaluated, and compare the robustness fluctuation caused by the number of the newly added sub-models.

Our experiment results reveal that the transformation-based ensemble indeed brings robustness to some extent, however, it cannot bring the expected strong robustness as shown previously in~\cite{meng2020ensembles}.
In addition, our results indicate that the gained robustness by transformation-based ensemble is mainly from the \itrans s instead of the ensemble of a number of sub-models. 
Specifically, for ensemble based on \rtrans s, the transferability of PGD-generated AEs among sub-models trained on data after different transformation reaches 59.71\% on average. 
As a consequence, in TAA, the AEs generated by PGD against one single sub-model can lower the classification accuracy of the ensemble to 32.64\% on average.
PAA attack can compromise the accuracy of the ensemble based on \rtrans s further. 
Even optimisation-based attacks, which show little harm to the ensemble in TAA, can successfully lower the accuracy of the ensemble to around 70\% using C\&W as the base attack algorithm and 20\% using DeepFool as the base attack algorithm.
For ensembles based on \itrans s, taking the PGD attack as an example, our results show that the ensemble accuracy, regardless of the used ensemble strategy, is only around 50\%, which is around 10\% higher than the ensemble based on \rtrans s.
In addition, the number of sub-models in an ensemble based on \itrans s does not affect the robustness against AEs much.

In summary, we made the following contributions.
\begin{itemize}
\item We analyse the transferability of AEs among the models trained on data after \rtrans s and evaluate the ensemble based on such models using the proposed TAA.
The experiment results show that the transferability does exist among the models performing the same classification task, even when they are trained on the data that have been pre-processed by different \rtrans s. 
And, the ensemble with only such sub-models is vulnerable to AEs generated against any of its sub-models regardless of its overall ensemble strategy.

\item Instead of targeting only one sub-model as in TAA, we further propose an adaptive and bounded PAA by aggregating the adversarial perturbations generated against all the ensemble sub-models using various aggregation strategies.
PAA adopts a dissimilarity score to bound the generated adversarial perturbations, which helps to generate more stealthy AEs.
The experiment results show that ensemble with only sub-models trained on data after \rtrans \ is vulnerable to PAA.
Those base attack algorithms, under which the reversible ensemble has shown strong robustness in TAA experiments, such as C\&W and DeepFool, also successfully compromise the ensemble accuracy largely in PAA experiments.

\item For the ensemble of sub-models trained on data after \itrans s, we evaluate its robustness using transferability-based attacks~\cite{meng2020ensembles}. 
We design experiments to identify the contribution to the robustness from \rtrans s and \itrans s, respectively.
The results indicate that 1) generally the transformation-based ensemble can provide certain robustness but cannot reach its debut robustness; 2) the gained robustness is from the \itrans \ instead of the ensemble of a number of sub-models; 3) the number of sub-models is not a key factor in gaining robustness, i.e., it cannot gain more robustness by incorporating more such sub-models.

\end{itemize}

\section{Background}\label{sec:background}
\subsection{Evasion Attacks}\label{subsec:evasion-attack}

A classification task can be represented as $y_{pred} = f_\theta(x)$, where a classifier $f$ with parameters $\theta$ takes input $x$ and outputs the prediction $y_{pred}$ for $x$. 
In the evasion attack, an AE $x^*$ is an input sample maliciously crafted to mislead the victim classifier to an erroneous output $y^* = f_\theta(x^*)$.
$x^*$ is close to $x$ with respect to some distance measurement $d_{L_p}$ such as $L_1, L_2, L_{\infty}$ norms, that is $d_{L_p}(x,x^*)<\epsilon$, so that the perturbation is imperceptible. 
If the erroneous output $y^*$ is unspecified as long as $y^*\neq y_{pred}$, the attack is named untargeted attack.
We focus on untargeted attacks in this paper, while the analysis can be easily extended to targeted attack. 

The evasion attacks can be categorised into the black-box attack and white-box attack.
In the black-box attack, the attacker has no knowledge of the internal information about the model.
The attacker can only perform legitimate operations such as querying using her/his data and observing the model output.
In the white-box setting, the adversary knows all about the model, including the model structure and parameter weights.
There have been many methods proposed for generating AEs in both black-box setting and white-box setting.

We introduce four classical white-box AE generation methods used in this paper, i.e., FGSM (Fast Gradient Sign Method)~\cite{fgsm}, PGD (Projected Gradient Descent)~\cite{madry2017towards}, C\&W~\cite{carlini2017towards}, and DeepFool~\cite{moosavi2016deepfool}.

FGSM~\cite{fgsm} is a classical gradient-based attack.
It updates the adversarial perturbations by the gradient of adversary's objective with respect to the input in the following way.
\[x^*=x+\epsilon * sign(\nabla_{x}L(f_{\theta}(x),y))\]
where $L$ is the objective function for trained target classifier $f_{\theta}$, $\epsilon$ is the perturbation budget.

PGD~\cite{madry2017towards} is an iterative variant of FGSM.
It is also a representative of various iterative methods, such as Basic Iterative Method \cite{BIM} and Momentum Iterative Method \cite{dong2018boosting}, due to its superior attack effect \cite{mao2019metric}.
PGD updates perturbation gradually till the perturbation budget exceeded or the target model successfully fooled.
During every iteration, instead of taking a big step at the size of budget $\epsilon$, PGD updates perturbations in a
gentle way with step size $\alpha < \epsilon$.
As pixel values for images are bounded, PGD projects generated adversarial perturbations into the feasible domain before entering the next iteration.

C\&W~\cite{carlini2017towards} represents another type of generating methods that treat the procedure of finding adversarial examples as an optimisation problem.
Instead of solving an optimisation problem with distance constraints, C\&W attack applies the change of variable method to ensure the generated AEs are in a feasible domain.
Also, to keep AEs stealthy and effective, a weighted sum of $L_p$ distance and attacker's loss is included in the optimisation terms.
The $L_2$ bounded C\&W attack is to solve the following optimisation problem.
\[w = argmin_{w^*\in R}||\frac{1}{2}(\tanh(w^*)+1)-x||_2^2 + c \cdot f\left(\frac{1}{2}(\tanh(w^*)+1)\right)\]
where $c$ is a constant found by lattice search.

DeepFool~\cite{moosavi2016deepfool} is another optimisation-based method but with a closed form representation of
minimum distance to the closest locally linear approximation of target classifier's decision boundaries as
\begin{align*}
    \hat{l} \leftarrow argmin_{k \neq \hat{k}(x)} \frac{|f'_{k}|}{||w'_{k}||_2}
\end{align*}
where $k \in K$ is one of all possible classes, $x$ is the original benign example, and
\begin{gather*}
    w'_{k} \leftarrow \nabla f_{k}(x_i) - \nabla f_{\hat{k}(x)}(x_i) \\
    f'_{k} \leftarrow f_{k}(x_i) - f_{\hat{k}(x)}(x_i)
\end{gather*}
DeepFool performs the above operations to update the current generated AE $x_i$ at step $i$ iteratively till it successfully fools the victim classifier.

\subsection{Transferability}\label{subsec:transferability}
Experiments in previous research show that AEs generated by attack algorithms targeting one specific model may successfully fool other models~\cite{ilyas2018black}\cite{demontis2019adversarial}.
This type of phenomenon is called transferability.
Demontis et al.~\cite{demontis2019adversarial} proposed to represent transferability numerically as the value of target classifier's objective function given the perturbed example and its ground truth label.

The transferable nature of AEs has been shown to be able to assist attackers to fool a well-protected model by generating adversarial examples on a surrogate model trained with only black-box queries~\cite{papernot2017practical}.
In an ensemble of many sub-models, it is expected that the transferability of AEs among the sub-models should be low.
In this way, the ensemble can still make the correct decision with a high probability since not all of the sub-models are fooled at the same time. 
However, the transferability of AEs generated on one sub-model may differ from that on another sub-model.
Hence, an attacker may be able to generate adversarial examples on one ensemble member, i.e., the sub-model whose AEs are with high transferability, and successfully transfer its AEs to other ensemble members, eventually fooling the whole ensemble model.

\subsection{Transformation-based Ensemble Defence}
\label{subsec:background:transformation}

Empirical evidence has shown that adversarial examples are sensitive to image transformations caused by the camera when performing attacks in the physical world~\cite{BIM}.
Taking pictures of successful adversarial images as input to the target classifiers, the attack success rate drops from nearly 100\% to almost zero.
The AE's high sensitivity to image transformation is also used in the detection of AE~\cite{xu2017feature}.
By comparing the softmax probability vectors across the outputs of the classifier trained on the original data and the classifier trained on the transformed data (i.e., colour-depth-reduction and spatially pixel smoothing), it can detect the existence of AE which is indicated by a high distance score.

\begin{figure}
    \centering
    \includegraphics[width=\columnwidth]{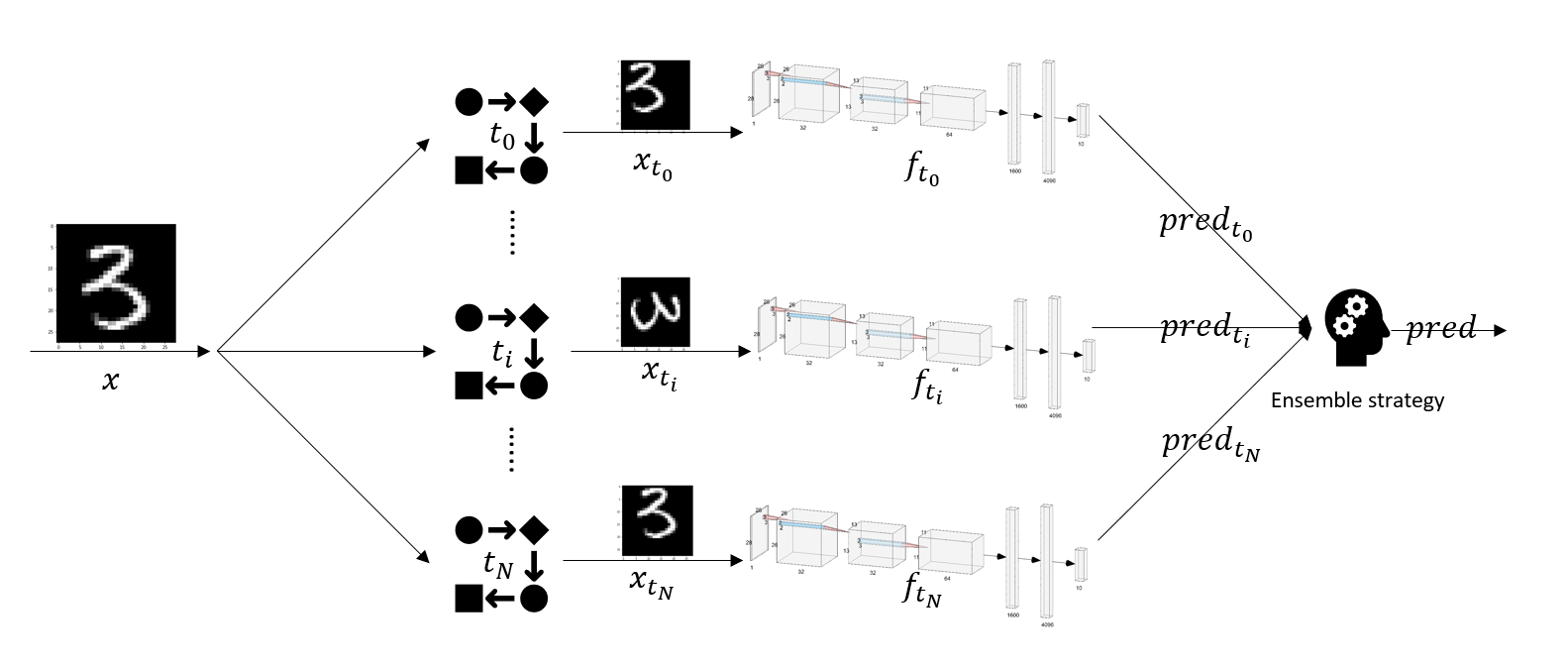}
    \caption{Illustration of a Transformation-Based Ensemble.}
    \label{fig:athena_demo}
\end{figure}

A very recent work proposes a robust ensemble with sub-models on 72 input transformations for image classification task~\cite{meng2020ensembles}. 
As shown in Figure~\ref{fig:athena_demo}, each sub-model is trained on the training data that have been gone through one specific transformation.
The 72 transformations are listed in Table~\ref{tab:trans_tab}.
We categorise them into reversible transformations and irreversible transformations.

\begin{table}[hbt!]
\centering
\caption{Transformations List}
\label{tab:trans_tab}
\resizebox{\columnwidth}{!}{%
\begin{tabular}{|c|l|}
\hline
\textbf{Category} & \multicolumn{1}{c|}{\textbf{Transformation Details}} \\ \hline
\multirow{3}{*}{\textbf{\begin{tabular}[c]{@{}c@{}}Reversible\\ Transformations\end{tabular}}} & Shift (left, right, top left, top right, bottom left, bottom right) \\ \cline{2-2} 
 & Flip (horizontal, vertical, both) \\ \cline{2-2} 
 & Rotation (90°,180°,270°) \\ \hline
\multirow{12}{*}{\textbf{\begin{tabular}[c]{@{}c@{}}Irreversible\\ Transformations\end{tabular}}} & Cartoonify (4 styles) \\ \cline{2-2} 
 & Affine (Compress / Stretch vertically and/or horizontally) \\ \cline{2-2} 
 & Denoise (nl means, nl means fast, tv menas, tv chambolle, wavelet) \\ \cline{2-2} 
 & Morphology (erosion, dilation, opening, closing, gradient) \\ \cline{2-2} 
 & Noise (gaussian, localvar, pepper, poison, salt, salt\&pepper) \\ \cline{2-2} 
 & Augmentation (feature-wise / sample-wise std normalization) \\ \cline{2-2} 
 & Segmentation (color-based) \\ \cline{2-2} 
 & Quantization (4 clusters, 8 clusters) \\ \cline{2-2} 
 & Distortion (x-axis, y-axis) \\ \cline{2-2} 
 & \begin{tabular}[c]{@{}l@{}}Filter (entropy, gaussian, maximum, median, minimum, Prewitt,\\ rank, Scharr, Roberts, Sobel)\end{tabular} \\ \cline{2-2} 
 & Compress (80\%, 50\%, 30\%, 10\% for jpeg, 1, 5, 8 for png) \\ \cline{2-2} 
 & Geometric (iradon, iradon sart and swirl) \\ \hline
\end{tabular}%
}
\end{table}

In the inference phase, the ensemble takes an original image $x$ as input. 
For each sub-model $f_{t_i}, i \in [0,71]$, the original $x$ is transformed by $t_i$ to $x_{t_i}$ before being processed by $f_{t_i}$.
The decisions of all the sub-models, in the form of predicted probabilities or logits from the last layer, can be aggregated using one of the 5 ensemble strategies, i.e., Random Defence (RD), Majority Voting (MV), Top 2 Majority Voting (T2MV), Average Probability (AVEP) and Average Logits (AVEL).
RD outputs the prediction from a random sub-model.
MV determines the output label that is agreed upon by most sub-models.
T2MV performs MV among labels associated with the top two probabilities predicted by each sub-model.
AVEP predicts based on the average predicted probabilities of all sub-models.
AVEL is based on average logits of all sub-models.

One thing worth noting is that not all of the 72 transformations can be reversed as shown in Table~\ref{tab:trans_tab}. 
Transformations that cannot be properly reversed may cause difficulties for the AE generation algorithms to cast the perturbation on $x_{t_i}$ to $x$.
Though the ensemble is claimed to be robust against AEs generated by several classical AE generation algorithms, we would like to find out \textit{where exactly the robustness comes from in the transformation-based ensemble?}

\section{Methodology}\label{sec:attack-methodology}


We observe that a transformation-based ensemble can include both reversible and irreversible transformations.
For ensemble based on reversible transformations, the adversarial perturbations are able to project back to the original image before the transformation.
However, for ensemble based on irreversible transformations, it is impossible to do so. 
Hence, it is necessary to treat them differently so as to procedure scientific analysis results.

This section mainly proposes two adaptive attacks: the TAA and PAA for the reversible ensembles, and then explains the core design of how we evaluate ensembles based on reversible and irreversible transformations, respectively.

\subsection{Transferability-based Adaptive Attack}\label{TAA}

Previous researches~\cite{papernot2017practical}\cite{cheng2019improving} have demonstrated that adversarial examples generated by attack algorithms targeting one specific model may successfully fool another model with similar functionality. 
In the transformation-based ensemble, each model is trained on the same training data but after different transformation. 
We conjecture that the transferability of AEs may still exist among these models since they are performing the same classification task.

The idea of Transferability-based Adaptive Attack (TAA) comes from Liebig's Law of the minimum in agricultural science.
It originally states that the growth is indicated not by the total resources but by the scarcest resource~\cite{Ebelhar2008}.
In the case of an ensemble, the robustness of the model to some extent is determined by the transferability of AEs generated by a single model. 
Adversarial examples generated on one ensemble member may transfer to other ensemble members, but the transferability rate differs as the victim model which the AEs are generated on differs.
Lower transferability indicates a lower probability that the entire ensemble is compromised at the same time.
Therefore, thinking from the attackers' angle, the model whose AEs have the highest transferability among all the ensemble sub-models is the scarcest resource they are looking for, i.e., the ``weakest'' sub-model in the ensemble.  
AEs generated against the ``weakest'' ensemble member may have a higher probability of successfully fooling the whole ensemble.

TAA, as shown in Algorithm~\ref{alg:TAA-algorithm}, is a naive attack in which adversarial examples are generated on one target ensemble member, whose AEs have the highest transferability, trying to fool the overall ensemble.
The target sub-model is selected according to the transferability ranking algorithm as shown in Algorithm~\ref{alg:trans_eval}.
Algorithm~\ref{alg:trans_eval} is to measure the transferability of the AEs generated on certain model by the accuracy of other sub-models on these AEs.
A lower accuracy indicates a larger number of models the generated AEs have fooled, and hence the AEs generated on that model have a higher transferability. 
This selection needs to be done only once in advance, and can be done on a much smaller test dataset $(\chi ^*, \gamma ^*)$, e.g., randomly selected 100 samples.

TAA is a basic algorithm to evaluate the robustness of an ensemble solely based on reversible transformations.
The measurement of the transferability of AEs using Algorithm~\ref{alg:trans_eval} can be used to understand the robustness source of the ensemble, which is demonstrated later in Section~\ref{subsec: transferability-results}.

\begin{algorithm}[t]
    \caption{Transferability-based Adaptive Attack.}
    \label{alg:TAA-algorithm}
    \KwInput{Benign sample $x$, corresponding ground truth label $y$, specific attack algorithm $attacker$, transferability ranking $tr$}
    \KwOutput{Adversarial example $ x_{adv} $}
    $target$ $\leftarrow$ $tr$[0] \\
    $x_{adv} \leftarrow $ $x$  \\
    $x_{trans} \leftarrow  $ transform($x_{adv}$, $target$) \\
    $x_{tmp}$ $\leftarrow$ attacker.attack($x_{trans}$, $y$) \\
    $ x_{adv} \leftarrow$ reset($x_{tmp}$) \\
    return $x_{adv}$
\end{algorithm}

\begin{algorithm}[t]
    \caption{Transferability Ranking Algorithm}
    \label{alg:trans_eval}
    \KwInput{Benign samples $\chi$, ground truth labels $\gamma$, attack algorithm $attacker$, set of sub-models $S_{sub-models}$}
    \KwOutput{Transferability ranking $tr$}
    $(\chi ^*, \gamma ^*) \leftarrow $ shuffle$(\chi, \gamma)[0:100]$ \\
    $performance$ $\leftarrow$ $Empty\ Dict$ \\
    \ForEach{cand $\in$ $S_{sub-models}$}{
        $ \chi_{adv} \leftarrow \chi^* $  \\
        $ \chi_{trans} \leftarrow  $ transform($\chi_{adv}$, cand) \\
        $ \chi_{tmp}$ $\leftarrow$ attacker.attack($\chi_{trans}$) \\
        $ \chi_{adv} \leftarrow$ reset($\chi_{tmp}$) \\
        $performance[cand]$ = accuracy(cand.predict($\chi_{adv}), \gamma^*$) \\
    }
    $tr$ = sort\_by\_values\_ascending($performance$) \\
    return $tr$
\end{algorithm}

\subsection{Perturbation Aggregation Attack}\label{PAA}

As we are targeting an ensemble instead of one single model, it is natural to aggregate adversarial perturbations generated on all ensemble members to perform adversarial attacks.
Hence, we propose PAA as explained in Algorithm~\ref{alg:PAA-algorithm}.
In PAA, a greedy attacker will keep generating perturbations on sub-models that have not been fooled and aggregate those perturbations with certain aggregation strategy, until the perturbation becomes larger than a preset perturbation budget.

Different from attack algorithms targeting one single model which are bounded by different norms~\cite{fgsm}\cite{carlini2017towards}\cite{BIM}\cite{madry2017towards}, attacking an ensemble of more than one sub-model may also introduce perturbation several times larger when generating on each of them. 
Therefore, PAA needs a bound score to control the overall perturbation budget.
Following previous work~\cite{meng2020ensembles}, we use the normalised $L_2$ dissimilarity to measure the perturbation as shown in Equation~\ref{eq:dissimilarity}.

\begin{equation}
    dissimilarity(x_{adv}, x_{0})=\frac{||x_{adv}-x_{0}||_2}{||x_{0}||_2}
    \label{eq:dissimilarity}
\end{equation}

In the following subsections, we introduce four perturbation aggregation strategies that could be used in PAA, including maximum perturbation (MaxP), average perturbation (AvgP), the majority vote of perturbations (MVoteP) and the weighted sum of perturbations (WSumP). 
Theoretically, they have their own advantages and disadvantages, however, their actual performance needs to be experimentally evaluated.

\subsubsection{Maximum Perturbation (MaxP)}
MaxP aggregates generated perturbations by selecting the perturbation with the maximum absolute value feature-wise, i.e., pixel values in image.
By MaxP, it is guaranteed that the largest perturbations can be preserved in the aggregated perturbation.

\subsubsection{Average Perturbation (AvgP)}
Taking the average of perturbations is another straightforward solution to aggregate the adversarial perturbations from all sub-models. 
It can averagely approximate the perturbation level pixel-wise.
One possible drawback is that there is a possibility that the individual perturbations are averaged out, because the perturbations can be positive or negative.

\subsubsection{Majority Vote Perturbations (MVoteP)}
MVoteP uses the pixel-level perturbation agreed upon by most generated perturbations as the final perturbation.
Theoretically, for attack algorithms like FGSM where the perturbations generated on different models are discrete (i.e., $\epsilon$, $0$ and $-\epsilon$), MVoteP may perform well because the number of candidates to be voted is limited to three.
For other attack algorithms involving iterations, the perturbation may have various values.
In that case, the meaning of voting to a large number of candidates may end up like random choosing one perturbation.

\subsubsection{Weighted Sum of Perturbations (WSumP)}

Previous three aggregation mechanisms treat all ensemble members equally.
Since the transferability of adversarial examples generated on different models may be different, we design WSumP to use the weighted sum of all perturbations from sub-models according to their transferability ranking $tr$ which can be calculated using Algorithm~\ref{alg:trans_eval}.

WSumP aggregation assigns different weights to the perturbations according to their transferability rankings. 
We use the following set of weights that empirically demonstrate good performance.
For the perturbations generated on the top two sub-models in the transferability ranking list, WSumP assign $N-2$ to them, where $N$ is the total number of ensemble members.
For the perturbations generated on sub-models at the lowest 80\% positions in the transferability ranking list, WSumP assigns $0.8*N$.
The rest perturbations are assigned with $N-i$, where $i$ is the the sub-model's position in the transferability ranking list.
A soft-max operation is performed on the weights to keep the sum of those weights equals to one and $w_i$ is the weight for the corresponding perturbation $perturb_i$.

\begin{gather}
    pos_i \leftarrow clipping(N - i, 0.8 * N, N - 2)\\
    w_i \leftarrow \frac{\exp{pos_{i}}}{\sum{\exp{pos_{i}}}}
    \label{eq:wsp}
\end{gather}
\begin{equation}
    perturb=\sum_{i\in N}w_i * perturb_i
\end{equation}
\SetArgSty{textnormal}
\begin{algorithm}[t]
    \caption{Perturbation Aggregation Attack}
    \label{alg:PAA-algorithm}
    \KwInput{Benign sample $x$, corresponding ground truth label $y$, set of sub-models $S_{sub-models}$, maximum dissimilarity $max\_dis$, transferability ranking $tr$}
    \KwOutput{Adversarial example $ x_{adv} $}
    $ S_{candidates} \leftarrow S_{sub-models}$ \\
    $x_{tmp} \leftarrow x$  \\
    $dis \leftarrow 0$ \\
    \While{$dis \leq max\_dis\ \text{and} \ S_{cand}\ \text{is not} \ None$}{

        $x_{adv}$ $\leftarrow$ $x_{tmp}$ \\
        $perturb_{tmp}$ $\leftarrow$  $Empty \  Dict$ \\
        \ForEach{$cand$ in $S_{candidates}$}{
            $x_{trans}$ $\leftarrow$ \text{transform}($x_{adv}$, $cand$) \\
            $perturb_{tmp}$[$cand$] $\leftarrow$ reset(attacker.attack($x_{trans}$)) \\
        }
        $x_{tmp}$ $\leftarrow$ aggregate($perturb_{tmp}, tr$) \\
        \ForEach{$cand$ $\in$ $S_{candidates}$}{
            \If{$cand$.predict(transform($x_{adv}$, $cand$)) $\neq y$}{
                $S_{candidates}$.remove($cand$)
            }
        }
        $dis$ $\leftarrow$ dissimilarity($x_{tmp}, x$)
    } 
    return $x_{adv}$
\end{algorithm}

\subsection{Evaluating Ensemble Defence based on Reversible and Irreversible Transformation}
For an ensemble based on reversible transformations, we can use the proposed adaptive attack, i.e., TAA and PAA, to better evaluate the ensemble robustness. 
We can also measure the transferability of AEs among the sub-models. 
We expect the attack results can provide us some evidence whether the ensemble based only on reversible transformation can gain some robustness.

For an ensemble including irreversible transformations, TAA and PAA cannot be adopted because they require the adversarial perturbation to be projected back to the original input which is impossible when the irreversible transformation exists.
Hence, we plan to evaluate the ensemble using an ensemble evaluation method used in previous literature~\cite{meng2020ensembles} which is also based on the transferability of AEs.

We maintain the reversible transformation ensemble unchanged, whose performance has been evaluated in previous steps.
We randomly draw a number of irreversible sub-models and add them into the unchanged reversible transformation ensemble while strictly controlling the proportion of the numbers of reversible and irreversible sub-models.
We shall observe the influence on the ensemble robustness against AEs from the number of newly-added irreversible sub-models.
Furthermore, if an ensemble demonstrates robustness, we intend to further measure the impact of the number of ensemble members on the ensemble robustness.

\section{Experiment}\label{sec:experiment}
\subsection{Experiment Setup}\label{subsec:experiment-setup}

\begin{figure}[hbt!]
    \centering
    \includegraphics[width=0.95\columnwidth]{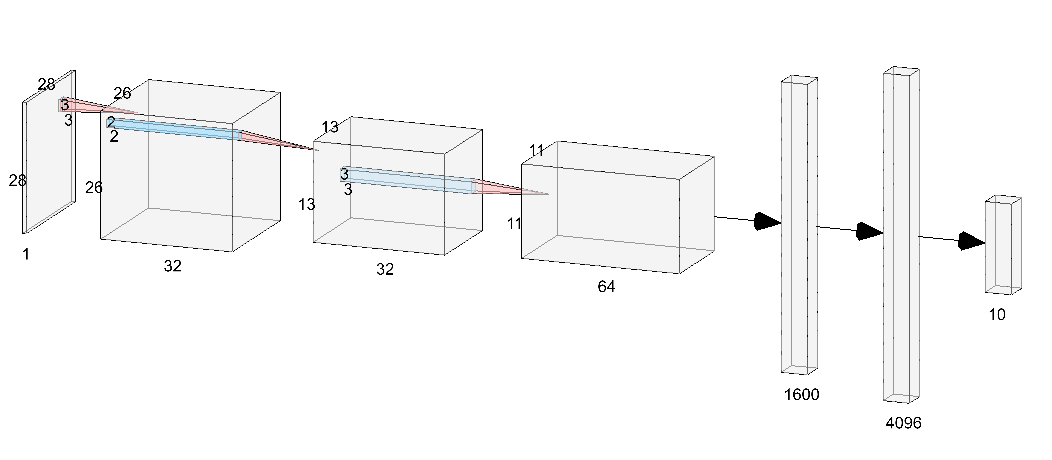}
    \caption{Structure of the Sub-Model.}
    \label{fig: mnist_model_structure}
\end{figure}

\textbf{Model.}
The sub-model structure is shown in Figure~\ref{fig: mnist_model_structure}, which consists of three convolutional layers followed by two fully connected layers and one soft-max layer~\cite{meng2020ensembles}. 
The transformation-based ensemble combines the sub-model in the way we introduced in Figure~\ref{fig:athena_demo} in Section~\ref{subsec:background:transformation}.
We put the 72 transformations (see Table~\ref{tab:trans_tab}) into our transformation pool, which includes 14 reversible transformations and 58 irreversible transformations. 
We construct the reversible transformation ensemble and irreversible transformation ensemble by drawing candidate transformation from this pool.
There are 5 ensemble strategies evaluated, i.e., Random Defence (RD), Majority Voting (MV), Top 2 Majority Voting (T2MV), Average Probabilities (AVEP), Average Logits (AVEL) (Detailed explanations can be found in Section~\ref{subsec:background:transformation}.).

\textbf{Dataset.} 
We evaluate the robustness of the ensemble on the MNIST dataset~\cite{lecun2010mnist}.
MNIST dataset contains 70,000 grey-scale images of hand-written digits.
We further split the original testing data (10,000 images) into validation and testing set at the ratio of 1:1.
Early-stopping technique based on cross-entropy loss on validation dataset is adopted to prevent models from over-fitting during training.

\textbf{Base attack algorithms used in TAA and PAA.}
We assume that the attacker has full knowledge of the ensemble including the transformation pre-processing, the model parameters, and the ensemble strategy. 
We utilise 4 representative attack algorithms in our experiments, i.e., FGSM, PGD, C\&W, and DeepFool.
We set the parameters for these algorithms in an overkill way~\cite{abbasi2017robustness}\cite{he2017adversarial}\cite{meng2020ensembles} as the findings in recent work by Tram{\`e}r et al~\cite{Carlini} indicate that proper attack parameters and sufficient iterations are essential for an attack to converge so as to provide convincing evaluation results.
The parameters are set as shown in Table~\ref{tab:attack_params}.

\begin{table}[hbt!]
\centering
\caption{Parameters for Attack Algorithms}
\label{tab:attack_params}
\resizebox{0.9\columnwidth}{!}{%
\begin{tabular}{|c|c|c|}
\hline
\textbf{\begin{tabular}[c]{@{}c@{}}Attack\\ Algorithm\end{tabular}} & \textbf{\begin{tabular}[c]{@{}c@{}}Parameters\\ in TAA\end{tabular}} & \textbf{\begin{tabular}[c]{@{}c@{}}Parameters\\ in PAA\end{tabular}} \\ \hline
\textbf{FGSM} & \begin{tabular}[c]{@{}c@{}}norm=$\infty$,\\ $\epsilon$=0.3\end{tabular} & \begin{tabular}[c]{@{}c@{}}norm=$\infty$,\\ $\epsilon$=0.05\end{tabular} \\ \hline
\textbf{PGD} & \begin{tabular}[c]{@{}c@{}}norm=$\infty$,\\ $\epsilon$=0.3,\\ maximum iterations=100\end{tabular} & \begin{tabular}[c]{@{}c@{}}norm=$\infty$,\\ $\epsilon$=0.05,\\ maximum iterations=250\end{tabular} \\ \hline
\textbf{C\&W} & \multicolumn{2}{c|}{\begin{tabular}[c]{@{}c@{}}norm=2,\\ binary search steps=10,\\ initial constant=0.01,\\ learning rate=0.01,\\ maximum iterations=100\end{tabular}} \\ \hline
\textbf{DeepFool} & \multicolumn{2}{c|}{\begin{tabular}[c]{@{}c@{}}norm=2,\\ maximum iterations=100,\\ overshoot=1e-6,\\ number of candidates=3\end{tabular}} \\ \hline
\end{tabular}%
}
\end{table}

\subsection{Transferability among Models Trained on Differently Transformed Training Data}\label{subsec: transferability-results}

\begin{table*}[t]
\centering
\caption{Transferability Rate of Adversarial Examples Generated on Different Sub-Models}
\label{tab:trans_result_mnist}
\resizebox{\textwidth}{!}{%
\begin{tabular}{|c|c|c|c|c|c|c|c|c|c|c|c|c|c|c|c|c|c|c|}
\hline
\textbf{\begin{tabular}[c]{@{}c@{}}Victim\\ Model\end{tabular}} & \textbf{\begin{tabular}[c]{@{}c@{}}Attack \\ Method\end{tabular}} & \textbf{\begin{tabular}[c]{@{}c@{}}no. \\ of AEs\end{tabular}} & \textbf{original} & \textbf{\begin{tabular}[c]{@{}c@{}}flip\\ both\end{tabular}} & \textbf{\begin{tabular}[c]{@{}c@{}}flip\\ horizontal\end{tabular}} & \textbf{\begin{tabular}[c]{@{}c@{}}flip\\ vertical\end{tabular}} & \textbf{\begin{tabular}[c]{@{}c@{}}rotate\\ 180\end{tabular}} & \textbf{\begin{tabular}[c]{@{}c@{}}rotate\\ 270\end{tabular}} & \textbf{\begin{tabular}[c]{@{}c@{}}rotate\\ 90\end{tabular}} & \textbf{\begin{tabular}[c]{@{}c@{}}shift\\ bottom\\ left\end{tabular}} & \textbf{\begin{tabular}[c]{@{}c@{}}shift\\ bottom\\ right\end{tabular}} & \textbf{\begin{tabular}[c]{@{}c@{}}shift\\ down\end{tabular}} & \textbf{\begin{tabular}[c]{@{}c@{}}shift\\ left\end{tabular}} & \textbf{\begin{tabular}[c]{@{}c@{}}shift\\ right\end{tabular}} & \textbf{\begin{tabular}[c]{@{}c@{}}shift\\ top\\ left\end{tabular}} & \textbf{\begin{tabular}[c]{@{}c@{}}shift\\ top\\ right\end{tabular}} & \textbf{\begin{tabular}[c]{@{}c@{}}shift\\ up\end{tabular}} & \textbf{Average} \\ \hline
\multirow{4}{*}{\textbf{origianl}} & \textbf{FGSM} & 3538 & 100.00\% & 54.61\% & 53.82\% & 50.59\% & 47.09\% & 38.75\% & 54.04\% & \textbf{59.61\%} & 44.52\% & 47.23\% & 56.11\% & 54.15\% & 39.15\% & 48.53\% & 45.20\% & 49.53\% \\ \cline{2-19} 
 & \textbf{PGD} & 4971 & 100.00\% & 75.32\% & 75.66\% & 72.84\% & 57.23\% & 57.33\% & 74.31\% & 69.66\% & 59.34\% & 65.84\% & \textbf{76.18\%} & 67.65\% & 48.78\% & 57.80\% & 64.43\% & 65.88\% \\ \cline{2-19} 
 & \textbf{C\&W} & 2303 & 100.00\% & 5.86\% & 9.03\% & 6.64\% & 6.08\% & 5.73\% & 8.03\% & \textbf{11.33\%} & 9.73\% & 8.94\% & 9.29\% & 10.81\% & 4.56\% & 7.08\% & 8.42\% & 7.97\% \\ \cline{2-19} 
 & \textbf{Deepfool} & 4844 & 100.00\% & 1.82\% & 2.31\% & 1.61\% & 1.51\% & 2.11\% & 1.32\% & \textbf{4.79\%} & 2.89\% & 2.75\% & 2.39\% & 2.29\% & 1.30\% & 3.53\% & 1.69\% & 2.31\% \\ \hline
\multirow{4}{*}{\textbf{\begin{tabular}[c]{@{}c@{}}flip\\ both\end{tabular}}} & \textbf{FGSM} & 3222 & 35.29\% & 100.00\% & 52.27\% & 49.50\% & 45.47\% & 37.96\% & 45.78\% & \textbf{58.75\%} & 41.37\% & 46.62\% & 48.57\% & 51.49\% & 37.21\% & 51.27\% & 37.31\% & 45.63\% \\ \cline{2-19} 
 & \textbf{PGD} & 4986 & 53.39\% & 100.00\% & \textbf{73.14\%} & 70.82\% & 60.55\% & 47.81\% & 65.50\% & 67.49\% & 54.21\% & 57.28\% & 69.33\% & 60.09\% & 46.23\% & 59.69\% & 50.82\% & 59.74\% \\ \cline{2-19} 
 & \textbf{C\&W} & 2742 & 6.20\% & 100.00\% & 8.68\% & 4.78\% & 7.77\% & 6.31\% & 6.71\% & \textbf{11.01\%} & 8.53\% & 7.59\% & 9.01\% & 8.97\% & 5.36\% & 9.30\% & 8.75\% & 7.78\% \\ \cline{2-19} 
 & \textbf{Deepfool} & 4811 & 2.45\% & 100.00\% & 1.75\% & 1.16\% & 1.60\% & 1.62\% & 1.27\% & \textbf{4.14\%} & 2.45\% & 1.97\% & 2.37\% & 1.89\% & 1.70\% & 2.74\% & 1.23\% & 2.03\% \\ \hline
\multirow{4}{*}{\textbf{\begin{tabular}[c]{@{}c@{}}flip\\ hori-\\ zontal\end{tabular}}} & \textbf{FGSM} & 3242 & 33.13\% & 45.43\% & 100.00\% & 41.64\% & 38.80\% & 28.22\% & 44.23\% & \textbf{53.27\%} & 39.39\% & 45.10\% & 45.56\% & 51.02\% & 33.50\% & 44.39\% & 30.54\% & 41.02\% \\ \cline{2-19} 
 & \textbf{PGD} & 4974 & 58.58\% & 72.34\% & 100.00\% & 67.23\% & 54.06\% & 40.57\% & \textbf{73.66\%} & 65.48\% & 58.85\% & 64.86\% & 70.71\% & 70.39\% & 46.86\% & 57.94\% & 61.74\% & 61.66\% \\ \cline{2-19} 
 & \textbf{C\&W} & 3246 & 4.65\% & 4.25\% & 100.00\% & 3.45\% & 3.82\% & 3.76\% & 5.85\% & \textbf{8.19\%} & 6.56\% & 5.55\% & 6.10\% & 7.05\% & 3.45\% & 5.36\% & 5.82\% & 5.28\% \\ \cline{2-19} 
 & \textbf{Deepfool} & 4824 & 1.64\% & 1.49\% & 100.00\% & 1.33\% & 1.29\% & 1.33\% & 1.18\% & \textbf{3.25\%} & 2.43\% & 1.49\% & 1.68\% & 1.70\% & 1.43\% & 2.16\% & 1.20\% & 1.69\% \\ \hline
\multirow{4}{*}{\textbf{\begin{tabular}[c]{@{}c@{}}flip\\ vertical\end{tabular}}} & \textbf{FGSM} & 3120 & 37.12\% & 56.63\% & 50.83\% & 100.00\% & 44.42\% & 37.88\% & 48.53\% & \textbf{61.31\%} & 47.98\% & 53.43\% & 52.60\% & 54.71\% & 36.63\% & 54.46\% & 35.26\% & 47.99\% \\ \cline{2-19} 
 & \textbf{PGD} & 4982 & 59.98\% & \textbf{76.80\%} & 72.02\% & 100.00\% & 55.52\% & 50.52\% & 69.95\% & 67.84\% & 60.26\% & 68.59\% & 74.55\% & 62.51\% & 46.95\% & 63.33\% & 56.70\% & 63.25\% \\ \cline{2-19} 
 & \textbf{C\&W} & 2555 & 7.79\% & 5.75\% & 8.57\% & 100.00\% & 7.51\% & 5.75\% & 8.10\% & \textbf{11.86\%} & 8.57\% & 8.77\% & 10.68\% & 9.43\% & 4.66\% & 8.96\% & 8.26\% & 8.19\% \\ \cline{2-19} 
 & \textbf{Deepfool} & 4833 & 2.46\% & 1.76\% & 2.13\% & 100.00\% & 1.74\% & 1.53\% & 1.43\% & \textbf{3.81\%} & 2.92\% & 2.40\% & 2.79\% & 1.97\% & 1.72\% & 3.23\% & 1.14\% & 2.22\% \\ \hline
\multirow{4}{*}{\textbf{\begin{tabular}[c]{@{}c@{}}rotate\\ 180\end{tabular}}} & \textbf{FGSM} & 3673 & 27.61\% & 44.49\% & 40.29\% & 36.78\% & 100.00\% & 35.94\% & 40.10\% & \textbf{56.47\%} & 36.13\% & 39.97\% & 41.66\% & 46.53\% & 33.38\% & 48.46\% & 38.91\% & 40.48\% \\ \cline{2-19} 
 & \textbf{PGD} & 4967 & 44.43\% & \textbf{66.68\%} & 60.90\% & 56.01\% & 100.00\% & 55.10\% & 59.57\% & 59.55\% & 46.77\% & 52.53\% & 55.33\% & 57.18\% & 44.15\% & 57.48\% & 50.75\% & 54.75\% \\ \cline{2-19} 
 & \textbf{C\&W} & 2704 & 4.62\% & 4.62\% & 5.29\% & 3.22\% & 100.00\% & 5.10\% & 5.10\% & \textbf{6.62\%} & 6.36\% & 5.92\% & 5.44\% & 6.32\% & 3.37\% & 5.10\% & 5.77\% & 5.20\% \\ \cline{2-19} 
 & \textbf{Deepfool} & 4819 & 1.74\% & 1.22\% & 1.56\% & 1.25\% & 100.00\% & 1.81\% & 1.08\% & \textbf{2.22\%} & 2.03\% & 1.45\% & 1.29\% & 1.51\% & 1.29\% & 1.89\% & 1.04\% & 1.53\% \\ \hline
\multirow{4}{*}{\textbf{\begin{tabular}[c]{@{}c@{}}rotate\\ 270\end{tabular}}} & \textbf{FGSM} & 3417 & 38.54\% & 50.28\% & 48.11\% & 44.28\% & 51.62\% & 100.00\% & 48.11\% & \textbf{62.54\%} & 44.54\% & 49.40\% & 49.96\% & 54.73\% & 43.69\% & 57.74\% & 49.84\% & 49.53\% \\ \cline{2-19} 
 & \textbf{PGD} & 4975 & 65.15\% & \textbf{73.65\%} & 68.30\% & 69.59\% & 76.60\% & 100.00\% & 64.22\% & 67.20\% & 58.23\% & 64.34\% & 69.43\% & 64.62\% & 54.99\% & 67.80\% & 73.63\% & 66.98\% \\ \cline{2-19} 
 & \textbf{C\&W} & 2038 & 10.84\% & 9.47\% & 10.01\% & 9.03\% & 12.27\% & 100.00\% & 9.27\% & 13.25\% & \textbf{10.89\%} & 9.18\% & 10.40\% & \textbf{11.73\%} & 6.43\% & 10.11\% & 11.53\% & 10.31\% \\ \cline{2-19} 
 & \textbf{Deepfool} & 4815 & 4.63\% & 2.99\% & 3.18\% & 2.76\% & \textbf{3.39\%} & 100.00\% & 1.81\% & 4.03\% & \textbf{3.45\%} & \textbf{2.66\%} & 3.03\% & 3.05\% & 1.89\% & \textbf{3.07\%} & 2.22\% & 3.01\% \\ \hline
\multirow{4}{*}{\textbf{\begin{tabular}[c]{@{}c@{}}rotate\\ 90\end{tabular}}} & \textbf{FGSM} & 3322 & 35.07\% & 47.56\% & 48.62\% & 40.49\% & 43.11\% & 32.30\% & 100.00\% & \textbf{57.41\%} & 39.28\% & 43.32\% & 48.62\% & 49.25\% & 32.87\% & 50.00\% & 33.68\% & 42.97\% \\ \cline{2-19} 
 & \textbf{PGD} & 4979 & 59.43\% & 65.92\% & \textbf{74.25\%} & 65.47\% & 55.19\% & 40.73\% & 100.00\% & 66.98\% & 53.85\% & 59.71\% & 71.44\% & 59.67\% & 42.68\% & 53.32\% & 47.86\% & 58.32\% \\ \cline{2-19} 
 & \textbf{C\&W} & 3157 & 5.89\% & 4.28\% & 6.18\% & 3.93\% & 4.47\% & 2.47\% & 100.00\% & \textbf{7.00\%} & 6.65\% & 5.57\% & 5.04\% & 5.48\% & 2.50\% & 3.99\% & 4.56\% & 4.86\% \\ \cline{2-19} 
 & \textbf{Deepfool} & 4809 & 1.71\% & 1.37\% & 1.41\% & 1.16\% & 1.39\% & 1.06\% & 100.00\% & \textbf{2.14\%} & 1.95\% & 1.46\% & 1.37\% & 1.31\% & 0.87\% & 1.77\% & 0.87\% & 1.42\% \\ \hline
\multirow{4}{*}{\textbf{\begin{tabular}[c]{@{}c@{}}shift\\ bottom\\ left\end{tabular}}} & \textbf{FGSM} & 4014 & 32.39\% & 42.60\% & 42.10\% & 35.13\% & 36.07\% & 26.91\% & 42.97\% & 100.00\% & 45.39\% & \textbf{56.13\%} & 42.43\% & 46.81\% & 31.99\% & 48.56\% & 28.23\% & 39.84\% \\ \cline{2-19} 
 & \textbf{PGD} & 4958 & 47.16\% & 59.48\% & 59.04\% & 48.37\% & 43.18\% & 36.12\% & 61.34\% & 100.00\% & 63.05\% & \textbf{77.11\%} & 60.67\% & 56.62\% & 44.68\% & 53.21\% & 34.59\% & 53.19\% \\ \cline{2-19} 
 & \textbf{C\&W} & 2511 & 5.85\% & 4.54\% & 6.21\% & 4.26\% & 3.78\% & 3.42\% & 4.02\% & 100.00\% & \textbf{8.68\%} & 8.52\% & 5.50\% & 6.33\% & 3.07\% & 4.42\% & 3.50\% & 5.15\% \\ \cline{2-19} 
 & \textbf{Deepfool} & 4809 & 1.85\% & 1.35\% & 1.58\% & 1.16\% & 0.94\% & 1.08\% & 0.85\% & 100.00\% & \textbf{2.43\%} & 1.81\% & 1.43\% & 1.54\% & 0.89\% & 1.33\% & 1.08\% & 1.38\% \\ \hline
\multirow{4}{*}{\textbf{\begin{tabular}[c]{@{}c@{}}shift\\ bottom\\ right\end{tabular}}} & \textbf{FGSM} & 3980 & 24.25\% & 34.17\% & 32.86\% & 32.91\% & 32.29\% & 24.10\% & 33.32\% & \textbf{51.66\%} & 100.00\% & 46.98\% & 30.73\% & 51.48\% & 20.33\% & 45.48\% & 23.22\% & 34.55\% \\ \cline{2-19} 
 & \textbf{PGD} & 4961 & 40.88\% & 47.51\% & 51.66\% & 49.00\% & 36.30\% & 32.15\% & 48.16\% & 61.18\% & 100.00\% & \textbf{71.78\%} & 47.25\% & 64.99\% & 29.69\% & 54.83\% & 34.51\% & 47.85\% \\ \cline{2-19} 
 & \textbf{C\&W} & 2479 & 4.64\% & 3.51\% & 5.08\% & 3.67\% & 2.94\% & 2.99\% & 3.63\% & \textbf{9.08\%} & 100.00\% & 5.93\% & 4.44\% & 6.05\% & 3.03\% & 5.57\% & 3.47\% & 4.57\% \\ \cline{2-19} 
 & \textbf{Deepfool} & 4822 & 1.51\% & 0.83\% & 1.31\% & 0.97\% & 1.16\% & 0.95\% & 0.79\% & \textbf{2.86\%} & 100.00\% & 1.91\% & 1.20\% & 1.51\% & 0.97\% & 2.24\% & 0.71\% & 1.35\% \\ \hline
\multirow{4}{*}{\textbf{\begin{tabular}[c]{@{}c@{}}shift\\ down\end{tabular}}} & \textbf{FGSM} & 3418 & 27.30\% & 37.45\% & 43.86\% & 36.42\% & 38.71\% & 26.62\% & 40.78\% & \textbf{63.78\%} & 52.14\% & 100.00\% & 41.43\% & 49.15\% & 25.01\% & 42.83\% & 25.89\% & 39.38\% \\ \cline{2-19} 
 & \textbf{PGD} & 4972 & 46.70\% & 56.32\% & 62.99\% & 56.80\% & 44.83\% & 36.71\% & 57.64\% & 77.96\% & \textbf{73.83\%} & 100.00\% & 63.25\% & 61.10\% & 33.39\% & 51.11\% & 41.23\% & 54.56\% \\ \cline{2-19} 
 & \textbf{C\&W} & 2879 & 5.04\% & 3.65\% & 5.18\% & 3.40\% & 3.99\% & 3.89\% & 4.48\% & 10.49\% & \textbf{11.11\%} & 100.00\% & 5.90\% & 5.77\% & 2.78\% & 5.21\% & 4.62\% & 5.39\% \\ \cline{2-19} 
 & \textbf{Deepfool} & 4818 & 1.70\% & 1.20\% & 1.18\% & 1.04\% & 1.20\% & 1.39\% & 0.98\% & \textbf{3.59\%} & 3.18\% & 100.00\% & 1.52\% & 1.33\% & 1.02\% & 2.14\% & 0.91\% & 1.60\% \\ \hline
\multirow{4}{*}{\textbf{\begin{tabular}[c]{@{}c@{}}shift\\ left\end{tabular}}} & \textbf{FGSM} & 3184 & 37.91\% & 46.26\% & 47.02\% & 41.05\% & 39.60\% & 32.16\% & 47.39\% & \textbf{58.13\%} & 34.33\% & 44.79\% & 100.00\% & 47.36\% & 36.97\% & 46.14\% & 33.79\% & 42.35\% \\ \cline{2-19} 
 & \textbf{PGD} & 4971 & 59.65\% & \textbf{74.37\%} & 73.51\% & 71.56\% & 53.37\% & 44.84\% & 71.86\% & 68.24\% & 49.85\% & 64.55\% & 100.00\% & 65.76\% & 44.24\% & 55.62\% & 45.32\% & 60.19\% \\ \cline{2-19} 
 & \textbf{C\&W} & 3144 & 6.20\% & 4.96\% & 7.38\% & 5.38\% & 5.25\% & 2.99\% & 5.79\% & 8.33\% & 6.74\% & 7.25\% & 100.00\% & \textbf{8.43\%} & 3.34\% & 4.61\% & 5.18\% & 5.85\% \\ \cline{2-19} 
 & \textbf{Deepfool} & 4804 & 1.60\% & 1.33\% & 1.75\% & 1.31\% & 1.31\% & 1.12\% & 0.96\% & \textbf{2.75\%} & 1.96\% & 1.52\% & 100.00\% & 1.42\% & 0.94\% & 1.67\% & 0.96\% & 1.47\% \\ \hline
\multirow{4}{*}{\textbf{\begin{tabular}[c]{@{}c@{}}shift\\ right\end{tabular}}} & \textbf{FGSM} & 3655 & 28.34\% & 40.16\% & 44.49\% & 35.95\% & 35.90\% & 29.49\% & 35.16\% & 52.50\% & 44.65\% & 42.60\% & 38.58\% & 100.00\% & 30.07\% & \textbf{54.06\%} & 32.80\% & 38.91\% \\ \cline{2-19} 
 & \textbf{PGD} & 4967 & 44.01\% & 57.56\% & 65.67\% & 54.66\% & 43.47\% & 37.59\% & 49.63\% & 57.94\% & 62.23\% & 57.28\% & 60.04\% & 100.00\% & 38.74\% & \textbf{65.11\%} & 45.00\% & 52.78\% \\ \cline{2-19} 
 & \textbf{C\&W} & 2922 & 3.97\% & 2.46\% & 4.11\% & 2.94\% & 3.52\% & 2.46\% & 3.08\% & \textbf{6.30\%} & 5.99\% & 3.90\% & 4.86\% & 100.00\% & 2.74\% & 5.41\% & 4.76\% & 4.04\% \\ \cline{2-19} 
 & \textbf{Deepfool} & 4812 & 1.68\% & 0.83\% & 1.25\% & 1.00\% & 1.00\% & 1.12\% & 0.89\% & \textbf{2.45\%} & 2.29\% & 1.39\% & 1.18\% & 100.00\% & 0.96\% & 2.33\% & 1.06\% & 1.39\% \\ \hline
\multirow{4}{*}{\textbf{\begin{tabular}[c]{@{}c@{}}shift\\ top\\ left\end{tabular}}} & \textbf{FGSM} & 4225 & 48.78\% & 66.08\% & 60.59\% & 59.55\% & 61.54\% & 53.02\% & 60.14\% & \textbf{74.06\%} & 49.14\% & 56.36\% & 61.96\% & 63.36\% & 100.00\% & 69.73\% & 61.40\% & \textit{\textbf{60.41\%}} \\ \cline{2-19} 
 & \textbf{PGD} & 4974 & 70.71\% & 83.21\% & \textbf{82.59\%} & 80.56\% & 77.16\% & 68.09\% & 80.94\% & 80.68\% & 65.08\% & 69.18\% & 79.37\% & 76.72\% & 100.00\% & 79.25\% & 82.25\% & \textit{\textbf{76.84\%}} \\ \cline{2-19} 
 & \textbf{C\&W} & 1490 & 12.42\% & 12.68\% & 17.72\% & 9.80\% & 13.02\% & 9.40\% & 10.81\% & \textbf{19.19\%} & 15.10\% & 13.89\% & 13.76\% & 16.78\% & 100.00\% & 15.10\% & 17.65\% & \textit{\textbf{14.09\%}} \\ \cline{2-19} 
 & \textbf{Deepfool} & 4811 & 4.76\% & 5.24\% & 5.28\% & 3.66\% & 3.03\% & 3.43\% & 2.60\% & \textbf{6.63\%} & 4.99\% & 3.37\% & 4.18\% & 3.89\% & 100.00\% & 5.94\% & 3.93\% & \textit{\textbf{4.35\%}} \\ \hline
\multirow{4}{*}{\textbf{\begin{tabular}[c]{@{}c@{}}shift\\ top\\ right\end{tabular}}} & \textbf{FGSM} & 3883 & 29.15\% & 44.14\% & 41.95\% & 41.77\% & 41.10\% & 35.85\% & 38.58\% & \textbf{57.53\%} & 44.60\% & 42.96\% & 42.21\% & 56.97\% & 35.31\% & 100.00\% & 43.99\% & 42.58\% \\ \cline{2-19} 
 & \textbf{PGD} & 4971 & 46.29\% & 65.16\% & 58.34\% & 61.15\% & 49.35\% & 46.17\% & 55.20\% & 61.44\% & 59.93\% & 53.97\% & 57.51\% & \textbf{71.03\%} & 48.72\% & 100.00\% & 55.84\% & 56.44\% \\ \cline{2-19} 
 & \textbf{C\&W} & 2371 & 5.61\% & 4.47\% & 6.20\% & 5.10\% & 5.44\% & 4.72\% & 4.56\% & 8.10\% & 8.65\% & 7.51\% & 5.86\% & \textbf{8.94\%} & 4.81\% & 100.00\% & 5.44\% & 6.10\% \\ \cline{2-19} 
 & \textbf{Deepfool} & 4824 & 1.78\% & 1.37\% & 1.49\% & 1.55\% & 1.39\% & 1.16\% & 0.89\% & 1.97\% & \textbf{2.63\%} & 1.58\% & 1.24\% & 2.09\% & 1.16\% & 100.00\% & 1.35\% & 1.55\% \\ \hline
\multirow{4}{*}{\textbf{\begin{tabular}[c]{@{}c@{}}shift\\ up\end{tabular}}} & \textbf{FGSM} & 3214 & 32.17\% & 44.49\% & 47.70\% & 38.46\% & 43.03\% & 37.40\% & 41.54\% & 58.59\% & 38.67\% & 42.81\% & 43.62\% & 50.19\% & 39.73\% & \textbf{53.27\%} & 100.00\% & 43.69\% \\ \cline{2-19} 
 & \textbf{PGD} & 4980 & 59.46\% & 65.60\% & \textbf{75.22\%} & 66.55\% & 65.56\% & 58.33\% & 62.97\% & 62.43\% & 55.00\% & 60.80\% & 62.11\% & 64.28\% & 59.48\% & 66.59\% & 100.00\% & 63.17\% \\ \cline{2-19} 
 & \textbf{C\&W} & 2696 & 7.94\% & 6.75\% & 9.53\% & 5.90\% & 7.83\% & 5.38\% & 6.38\% & 11.20\% & 8.79\% & 8.72\% & 7.31\% & \textbf{11.28\%} & 5.71\% & 9.31\% & 100.00\% & 8.00\% \\ \cline{2-19} 
 & \textbf{Deepfool} & 4826 & 2.51\% & 1.78\% & 2.13\% & 1.41\% & 1.66\% & 1.89\% & 1.35\% & 2.90\% & 3.03\% & 1.91\% & 1.82\% & 2.38\% & 1.68\% & \textbf{3.32\%} & 100.00\% & 2.13\% \\ \hline
\end{tabular}%
}
\end{table*}

It has been observed that transferability of AEs exists among models performing the same classification task \cite{papernot2017practical}.
This experiment is designed to understand whether the transferability still exists among the models trained on the same data but after different transformations. 
It refers to reversible transformation here because the perturbation generated on input after irreversible transformation cannot be projected back to the original input, and hence it does not make much sense to evaluate its transferability when the entire image is disturbed by the irreversible transformation. 

First of all, we trained 15 models, one model of which is trained on the original MNIST training data records while the other 14 are trained on the same batch of data records but after the 14 reversible transformations as shown in Table~\ref{tab:trans_tab} with the label unchanged, respectively.
Then, we generate AEs with FGSM, PGD, C\&W and DeepFool on every model and evaluate their transferability rates.
The transferability rate of the AEs generated on one model is measured by calculating the rate of successful attacks on other models.
The results are demonstrated in Table~\ref{tab:trans_result_mnist}.
The transformation operations are used to indicate the model trained on the data after that transformation, while ``original'' means the model trained on the original data records.

It can be seen in Table~\ref{tab:trans_result_mnist} that AEs generated using FGSM and PGD generally have high transferability rates.
On average 49.53\% of AEs generated on the ``original'' model with FGSM can fool models trained on data after reversible transformations.
This number for PGD is even more significant, with on average 65.88\% of AEs generated on the ``original'' model using PGD can fool the other models while this number rises to as high as 76.84\% on AEs generated on ``shift top left'' model using PGD. 

Generally, for the AEs generated using FGSM and PGD on models trained on data after one specific transformation, they all have a high successful rate in fooling the models trained on other transformations.
This indicates that the transferability of AEs \textit{does exist} among models trained on data after reversible transformations.
Such models cannot block the AEs from transferring, meaning that those models may have similar decision boundaries even trained on data after different transformations.
It also implies that an ensemble of such models may not be able to improve the robustness as long as the transformation is reversible.
Since the ensemble strategy also plays an important role in an ensemble, this conjecture needs further validation which will be shown in Section~\ref{subsec:experiment:taa}.

Most importantly, it is observed that the transferability rate of AEs differs as the victim model differs.
For example, on average 76.18\% of AEs generated on the model trained on examples shifted to their top-left with PGD can transfer to other sub-models, while only 34.59\% of AEs generated with PGD on the model trained on examples shifted to their bottom left can transfer to other models.
Hence, it implies a good tactic for the attacker to fool the ensemble by generating AEs targeting the top sub-model in the transferability ranking list. 

Also, we notice that AEs generated with gradient-based attack algorithms (e.g., PGD and FGSM) are more transferable compared to those generated with optimisation-based attack algorithms (e.g., C\&W and DeepFool) while the number of successful AEs on their victim models are at the same scale.
Due to the facts that optimisation-based attacks like C\&W and DeepFool generate more subtle perturbations and more specific to the victim model, it is reasonable that the transferability rate of their generated perturbations is lower.

\subsection{TAA Experiment Results}
\label{subsec:experiment:taa}
As shown in Section~\ref{subsec: transferability-results}, despite the models are trained on data that have been gone through different reversible transformations, AEs generated on one model can actually transfer to other models for the same classification task at a high probability.
Therefore, the ensemble based on reversible transformations may face a potential threat that AEs generated on one of its sub-models may fool the entire ensemble. 
However, it still depends on the ensemble strategy adopted in the ensemble.
In an untargeted attack, AEs are generated to fool the victim sub-model no matter which label the erroneous output is.
For example, if the ensemble uses majority voting ensemble strategy, it is not sure whether the AE can fool the majority sub-models to output an agreed erroneous output.
Hence, this experiment is necessary to better evaluate the robustness of the ensemble based on reversible transformations under various ensemble strategies. 

The target ensemble is the ensemble of the 14 sub-models that are trained on the data after 14 reversible transformations.
The ``original'' model is not included in this ensemble.
We generate AEs on every sub-model using attack algorithms on the test dataset of 5000 examples.
Five ensemble strategies are measured, i.e., RD, MV, T2MV, AVEP, and AVEL.

In practice, an attacker can select to attack only the sub-model, whose AE has the highest transferability, to achieve a high success attack rate, which is the core idea of TAA.
Here, we show the full results instead of just the top $tr$ ranked one for readers to have a better overview and understanding of attack results on ensembles using different ensemble strategies when choosing different target sub-model.
The overall results in terms of the ensemble classification accuracy under such attacks are listed in Table~\ref{tab:TAA_results}.
The performance of the ensemble under no attack is shown as ``benign''. 

Under PGD-base TAA, the ``shift top left'' is at the top of the $tr$ list.
It can be seen from Table~\ref{tab:TAA_results} that attackers can lower the ensemble accuracy to around 15\% by generating AEs against this sub-model using PGD-based TAA. 
Generally, AEs generated by FGSM and PGD lower the ensemble accuracy more than that by C\&W and DeepFool, which is consistent with the transferability ranking experimental results. 

It is worth noting that the robustness from integrating different reversible transformation sub-models is fragile even when the attacker only knows the classification task and training data.
This can be seen from the Table~\ref{tab:TAA_results} where the AEs generated against the ``original'' model which is not even in the ensemble can lower the ensemble accuracy to around 30\%.

\begin{table}[hbt!]
\centering
\caption{Ensemble Accuracy on AEs Generated on Only One Sub-Model. The last column ``Dis'' is the normalised dissimilarity score defined in Equation~\ref{eq:dissimilarity}.}
\label{tab:TAA_results}
\resizebox{\columnwidth}{!}{%
\begin{tabular}{|c|c|c|c|c|c|c|c|c|}
\hline
\textbf{\begin{tabular}[c]{@{}c@{}}Target \\ Model\end{tabular}} & \textbf{\begin{tabular}[c]{@{}c@{}}Attack \\ Method\end{tabular}} & \textbf{RD} & \textbf{MV} & \textbf{T2MV} & \textbf{AP} & \textbf{AL} & \textbf{Average} & \textbf{Dis} \\ \hline
\multicolumn{2}{|c|}{\textbf{Benign}} & 99.36\% & 99.70\% & 96.48\% & 99.68\% & 99.68\% & 98.81\% & N.A. \\ \hline
\multirow{4}{*}{\textbf{original}} & \textbf{FGSM} & 58.68\% & 69.44\% & 66.84\% & 69.24\% & 68.52\% & 66.54\% & 0.5565 \\ \cline{2-9} 
 & \textbf{PGD} & 34.26\% & 33.24\% & 34.84\% & 31.54\% & 30.62\% & 32.90\% & 0.6077 \\ \cline{2-9} 
 & \textbf{C\&W} & 97.00\% & 98.58\% & 92.82\% & 98.86\% & 98.82\% & 97.22\% & 0.1089 \\ \cline{2-9} 
 & \textbf{Deepfool} & 97.90\% & 99.50\% & 92.30\% & 99.52\% & 99.62\% & 97.77\% & 0.1363 \\ \hline
\multirow{4}{*}{\textbf{\begin{tabular}[c]{@{}c@{}}flip \\ both\end{tabular}}} & \textbf{FGSM} & 67.82\% & 72.94\% & 66.68\% & 72.10\% & 70.90\% & 70.09\% & 0.5493 \\ \cline{2-9} 
 & \textbf{PGD} & 30.42\% & 38.54\% & 26.68\% & 37.02\% & 21.74\% & 30.88\% & 0.5952 \\ \cline{2-9} 
 & \textbf{C\&W} & 94.52\% & 98.42\% & 92.02\% & 98.52\% & 98.62\% & 96.42\% & 0.1344 \\ \cline{2-9} 
 & \textbf{Deepfool} & 97.16\% & 99.38\% & 91.60\% & 99.46\% & 99.54\% & 97.43\% & 0.1287 \\ \hline
\multirow{4}{*}{\textbf{\begin{tabular}[c]{@{}c@{}}flip\\ hori-\\ zontal\end{tabular}}} & \textbf{FGSM} & 69.00\% & 75.66\% & 69.54\% & 75.00\% & 73.00\% & 72.44\% & 0.5488 \\ \cline{2-9} 
 & \textbf{PGD} & 28.30\% & 35.94\% & 24.46\% & 34.00\% & 13.60\% & 27.26\% & 0.5940 \\ \cline{2-9} 
 & \textbf{C\&W} & 94.84\% & 98.64\% & 91.58\% & 98.70\% & 98.90\% & 96.53\% & 0.1508 \\ \cline{2-9} 
 & \textbf{Deepfool} & 98.72\% & 99.46\% & 90.34\% & 99.50\% & 99.62\% & 97.53\% & 0.1203 \\ \hline
\multirow{4}{*}{\textbf{\begin{tabular}[c]{@{}c@{}}flip \\ vertical\end{tabular}}} & \textbf{FGSM} & 63.92\% & 71.50\% & 66.96\% & 70.82\% & 69.28\% & 68.50\% & 0.5841 \\ \cline{2-9} 
 & \textbf{PGD} & 31.40\% & 35.82\% & 27.04\% & 33.30\% & 17.54\% & 29.02\% & 0.6202 \\ \cline{2-9} 
 & \textbf{C\&W} & \textbf{94.40\%} & 98.48\% & 91.54\% & 98.48\% & 98.52\% & 96.28\% & 0.1257 \\ \cline{2-9} 
 & \textbf{Deepfool} & 98.10\% & 99.40\% & 90.30\% & 99.50\% & 99.54\% & 97.37\% & 0.1366 \\ \hline
\multirow{4}{*}{\textbf{\begin{tabular}[c]{@{}c@{}}rotate\\ 180\end{tabular}}} & \textbf{FGSM} & 71.98\% & 74.62\% & 68.66\% & 73.42\% & 72.14\% & 72.16\% & 0.5423 \\ \cline{2-9} 
 & \textbf{PGD} & 42.46\% & 44.14\% & 32.30\% & 42.68\% & 28.34\% & 37.98\% & 0.5758 \\ \cline{2-9} 
 & \textbf{C\&W} & 96.70\% & 98.94\% & 92.34\% & 98.88\% & 99.06\% & 97.18\% & 0.1245 \\ \cline{2-9} 
 & \textbf{Deepfool} & 98.02\% & 99.62\% & 91.16\% & 99.64\% & 99.66\% & 97.62\% & 0.1284 \\ \hline
\multirow{4}{*}{\textbf{\begin{tabular}[c]{@{}c@{}}rotate\\ 270\end{tabular}}} & \textbf{FGSM} & 57.56\% & 68.04\% & 63.56\% & 67.94\% & 67.12\% & 64.84\% & 0.5486 \\ \cline{2-9} 
 & \textbf{PGD} & 26.48\% & 28.42\% & 19.88\% & 26.90\% & 17.40\% & 23.82\% & 0.5731 \\ \cline{2-9} 
 & \textbf{C\&W} & 95.34\% & 98.14\% & 91.70\% & 98.10\% & \textbf{98.18\%} & 96.29\% & 0.1043 \\ \cline{2-9} 
 & \textbf{Deepfool} & 97.98\% & \textbf{99.16\%} & 87.68\% & \textbf{99.22\%} & \textbf{99.32\%} & 96.67\% & 0.1463 \\ \hline
\multirow{4}{*}{\textbf{\begin{tabular}[c]{@{}c@{}}rotate\\ 90\end{tabular}}} & \textbf{FGSM} & 65.18\% & 74.04\% & 67.54\% & 73.22\% & 70.80\% & 70.16\% & 0.5785 \\ \cline{2-9} 
 & \textbf{PGD} & 50.92\% & 39.78\% & 27.22\% & 38.10\% & 18.72\% & 34.95\% & 0.5808 \\ \cline{2-9} 
 & \textbf{C\&W} & 96.98\% & 98.94\% & 92.22\% & 99.06\% & 98.98\% & 97.24\% & 0.1505 \\ \cline{2-9} 
 & \textbf{Deepfool} & 98.76\% & 99.60\% & 91.70\% & 99.64\% & 99.70\% & 97.88\% & 0.1216 \\ \hline
\multirow{4}{*}{\textbf{\begin{tabular}[c]{@{}c@{}}shift \\ bottom \\ left\end{tabular}}} & \textbf{FGSM} & 53.50\% & 73.74\% & 68.40\% & 72.50\% & 69.38\% & 67.50\% & 0.5864 \\ \cline{2-9} 
 & \textbf{PGD} & 39.26\% & 46.88\% & 35.18\% & 45.08\% & 30.70\% & 39.42\% & 0.5951 \\ \cline{2-9} 
 & \textbf{C\&W} & 94.98\% & 99.26\% & 92.96\% & 99.22\% & 99.20\% & 97.12\% & 0.1081 \\ \cline{2-9} 
 & \textbf{Deepfool} & 99.02\% & 99.62\% & 93.68\% & 99.66\% & 99.72\% & 98.34\% & 0.1190 \\ \hline
\multirow{4}{*}{\textbf{\begin{tabular}[c]{@{}c@{}}shift\\ bottom \\ right\end{tabular}}} & \textbf{FGSM} & 73.26\% & 77.12\% & 71.96\% & 77.22\% & 75.46\% & 75.00\% & 0.5160 \\ \cline{2-9} 
 & \textbf{PGD} & 38.50\% & 53.82\% & 40.34\% & 51.80\% & 40.44\% & 44.98\% & 0.5970 \\ \cline{2-9} 
 & \textbf{C\&W} & 98.08\% & 99.04\% & 93.20\% & 99.08\% & 99.20\% & 97.72\% & 0.1079 \\ \cline{2-9} 
 & \textbf{Deepfool} & 97.84\% & 99.62\% & 92.28\% & 99.64\% & 99.70\% & 97.82\% & 0.1146 \\ \hline
\multirow{4}{*}{\textbf{\begin{tabular}[c]{@{}c@{}}shift \\ down\end{tabular}}} & \textbf{FGSM} & 80.66\% & 76.82\% & 69.72\% & 76.16\% & 74.10\% & 75.49\% & 0.5391 \\ \cline{2-9} 
 & \textbf{PGD} & 26.76\% & 44.94\% & 32.22\% & 43.32\% & 28.06\% & 35.06\% & 0.6097 \\ \cline{2-9} 
 & \textbf{C\&W} & 96.54\% & 99.06\% & 92.10\% & 99.08\% & 99.16\% & 97.19\% & 0.1316 \\ \cline{2-9} 
 & \textbf{Deepfool} & 98.64\% & 99.54\% & 93.44\% & 99.64\% & 99.68\% & 98.19\% & 0.1183 \\ \hline
\multirow{4}{*}{\textbf{\begin{tabular}[c]{@{}c@{}}shift \\ left\end{tabular}}} & \textbf{FGSM} & 68.38\% & 76.06\% & 68.54\% & 75.30\% & 73.28\% & 72.31\% & 0.5346 \\ \cline{2-9} 
 & \textbf{PGD} & 28.90\% & 38.86\% & 26.58\% & 36.00\% & 18.70\% & 29.81\% & 0.5962 \\ \cline{2-9} 
 & \textbf{C\&W} & 96.06\% & 98.44\% & \textbf{89.98\%} & 98.58\% & 98.64\% & 96.34\% & 0.1491 \\ \cline{2-9} 
 & \textbf{Deepfool} & 98.56\% & 99.58\% & 92.16\% & 99.58\% & 99.58\% & 97.89\% & 0.1182 \\ \hline
\multirow{4}{*}{\textbf{\begin{tabular}[c]{@{}c@{}}shift \\ right\end{tabular}}} & \textbf{FGSM} & 70.76\% & 75.38\% & 69.04\% & 74.62\% & 71.82\% & 72.32\% & 0.5219 \\ \cline{2-9} 
 & \textbf{PGD} & 56.72\% & 46.88\% & 33.50\% & 45.32\% & 27.10\% & 41.90\% & 0.6094 \\ \cline{2-9} 
 & \textbf{C\&W} & 97.12\% & 99.16\% & 92.88\% & 99.20\% & 99.24\% & 97.52\% & 0.1288 \\ \cline{2-9} 
 & \textbf{Deepfool} & 97.40\% & 99.52\% & 92.42\% & 99.58\% & 99.58\% & 97.70\% & 0.1211 \\ \hline
\multirow{4}{*}{\textit{\textbf{\begin{tabular}[c]{@{}c@{}}shift \\ top \\ left\end{tabular}}}} & \textbf{FGSM} & \textbf{34.36\%} & \textbf{48.30\%} & \textbf{44.26\%} & \textbf{47.54\%} & \textbf{46.88\%} & \textbf{44.27\%} & 0.5610 \\ \cline{2-9} 
 & \textbf{PGD} & \textbf{24.08\%} & \textbf{16.74\%} & \textbf{11.76\%} & \textbf{15.38\%} & \textbf{10.54\%} & \textbf{15.70\%} & 0.6127 \\ \cline{2-9} 
 & \textbf{C\&W} & 95.18\% & \textbf{97.78\%} & 91.22\% & 98.02\% & 98.30\% & 96.10\% & 0.0798 \\ \cline{2-9} 
 & \textbf{Deepfool} & \textbf{93.98\%} & 99.28\% & \textbf{85.84\%} & 99.26\% & 99.36\% & \textbf{95.54\%} & 0.1617 \\ \hline
\multirow{4}{*}{\textbf{\begin{tabular}[c]{@{}c@{}}shift \\ top \\ right\end{tabular}}} & \textbf{FGSM} & 72.06\% & 71.04\% & 64.60\% & 70.26\% & 68.20\% & 69.23\% & 0.5271 \\ \cline{2-9} 
 & \textbf{PGD} & 42.72\% & 42.84\% & 29.26\% & 41.54\% & 27.24\% & 36.72\% & 0.6096 \\ \cline{2-9} 
 & \textbf{C\&W} & 96.48\% & 98.88\% & 92.42\% & 98.94\% & 98.94\% & 97.13\% & 0.1086 \\ \cline{2-9} 
 & \textbf{Deepfool} & 99.02\% & 99.42\% & 91.70\% & 99.54\% & 99.54\% & 97.84\% & 0.1263 \\ \hline
\multirow{4}{*}{\textbf{\begin{tabular}[c]{@{}c@{}}shift\\ up\end{tabular}}} & \textbf{FGSM} & 61.88\% & 75.40\% & 70.48\% & 74.78\% & 73.30\% & 71.17\% & 0.5464 \\ \cline{2-9} 
 & \textbf{PGD} & 33.24\% & 34.66\% & 25.56\% & 32.90\% & 19.62\% & 29.20\% & 0.6072 \\ \cline{2-9} 
 & \textbf{C\&W} & 94.80\% & 98.00\% & 90.56\% & \textbf{98.22\%} & 98.32\% & \textbf{95.98\%} & 0.1283 \\ \cline{2-9} 
 & \textbf{Deepfool} & 98.26\% & 99.50\% & 90.28\% & 99.48\% & 99.54\% & 97.41\% & 0.1316 \\ \hline
\end{tabular}%
}
\end{table}

\subsection{PAA Experiment Results}

\begin{table}
\centering
\caption{Ensemble Accuracy on AEs Generated by PAA}
\label{tab:PAA_results}
\resizebox{0.9\columnwidth}{!}{%
\begin{tabular}{|c|c|l|l|l|l|l|}
\hline
\multirow{3}{*}{\textbf{\begin{tabular}[c]{@{}c@{}}Attack\\ Algorithm\end{tabular}}} & \multirow{3}{*}{\textbf{\begin{tabular}[c]{@{}c@{}}Aggregation\\ Strategy\end{tabular}}} & \multicolumn{5}{c|}{\multirow{2}{*}{\textbf{Ensemble Strategy}}} \\
 &  & \multicolumn{5}{c|}{} \\ \cline{3-7} 
 &  & \multicolumn{1}{c|}{\textbf{RD}} & \multicolumn{1}{c|}{\textbf{MV}} & \multicolumn{1}{c|}{\textbf{T2MV}} & \multicolumn{1}{c|}{\textbf{AVEP}} & \multicolumn{1}{c|}{\textbf{AVEL}} \\ \hline
\multicolumn{2}{|c|}{\textbf{Benign}} & \multicolumn{1}{c|}{99.36\%} & \multicolumn{1}{c|}{99.70\%} & \multicolumn{1}{c|}{96.48\%} & \multicolumn{1}{c|}{99.68\%} & \multicolumn{1}{c|}{99.68\%} \\ \hline
\multirow{4}{*}{\textbf{FGSM}} & \textbf{MaxP} & \multicolumn{1}{c|}{94.19\%} & \multicolumn{1}{c|}{94.87\%} & \multicolumn{1}{c|}{88.68\%} & \multicolumn{1}{c|}{95.31\%} & \multicolumn{1}{c|}{95.51\%} \\ \cline{2-7} 
 & \textbf{AvgP} & \multicolumn{1}{c|}{21.25\%} & \multicolumn{1}{c|}{20.59\%} & \multicolumn{1}{c|}{39.42\%} & \multicolumn{1}{c|}{20.63\%} & \multicolumn{1}{c|}{22.30\%} \\ \cline{2-7} 
 & \textbf{MVoteP} & \multicolumn{1}{c|}{47.86\%} & \multicolumn{1}{c|}{63.68\%} & \multicolumn{1}{c|}{65.38\%} & \multicolumn{1}{c|}{63.66\%} & \multicolumn{1}{c|}{65.30\%} \\ \cline{2-7} 
 & \textbf{WSumP} & \multicolumn{1}{c|}{21.13\%} & \multicolumn{1}{c|}{21.90\%} & \multicolumn{1}{c|}{39.94\%} & \multicolumn{1}{c|}{21.96\%} & \multicolumn{1}{c|}{24.02\%} \\ \hline
\multirow{4}{*}{\textbf{PGD}} & \textbf{MaxP} & 84.11\% & 90.40\% & 86.82\% & 90.63\% & 91.31\% \\ \cline{2-7} 
 & \textbf{AvgP} & 22.66\% & \textbf{15.32\%} & 38.28\% & \textbf{15.28\%} & \textbf{17.25\%} \\ \cline{2-7} 
 & \textbf{MVoteP} & 62.00\% & 68.09\% & 71.11\% & 68.65\% & 70.21\% \\ \cline{2-7} 
 & \textbf{WSumP} & 17.41\% & 16.65\% & 38.36\% & 16.97\% & 19.23\% \\ \hline
\multirow{4}{*}{\textbf{C\&W}} & \textbf{MaxP} & 92.86\% & 91.84\% & 90.82\% & 91.84\% & 92.86\% \\ \cline{2-7} 
 & \textbf{AvgP} & \multicolumn{1}{c|}{52.04\%} & \multicolumn{1}{c|}{62.24\%} & \multicolumn{1}{c|}{72.45\%} & \multicolumn{1}{c|}{57.17\%} & \multicolumn{1}{c|}{56.12\%} \\ \cline{2-7} 
 & \textbf{MVoteP} & \multicolumn{1}{c|}{85.71\%} & \multicolumn{1}{c|}{92.86\%} & \multicolumn{1}{c|}{85.23\%} & \multicolumn{1}{c|}{94.90\%} & \multicolumn{1}{c|}{93.88\%} \\ \cline{2-7} 
 & \textbf{WSumP} & 65.31\% & 76.53\% & 75.51\% & 69.39\% & 67.35\% \\ \hline
\multirow{4}{*}{\textbf{DeepFool}} & \textbf{MaxP} & 39.88\% & 40.59\% & 54.35\% & 40.74\% & 42.82\% \\ \cline{2-7} 
 & \textbf{AvgP} & 25.51\% & 23.47\% & 32.68\% & 23.47\% & 21.43\% \\ \cline{2-7} 
 & \textbf{MVoteP} & \textbf{16.25\%} & 15.68\% & \textbf{31.01\%} & 16.84\% & 18.27\% \\ \cline{2-7} 
 & \textbf{WSumP} & 25.51\% & 17.35\% & 31.63\% & 19.39\% & 20.41\% \\ \hline
\end{tabular}%
}
\end{table}

Unlike TAA which depends on the transferable nature of AEs to attack an ensemble of models, PAA is more specifically designed for ensemble defence.
It utilises all possible perturbations generated on sub-models to perform an adversarial attack on an ensemble.

The stealthiness of AEs generated by PAA are bound by normalised dissimilarity (see Equation~\ref{eq:dissimilarity}).
The dissimilarity score is set to 0.3 in this experiment since we notice the average dissimilarity in Table~\ref{tab:TAA_results} is around this value.
The classification accuracy of ensemble under PAA-generated AEs on the test dataset is shown in Table~\ref{tab:PAA_results}.
Four perturbation aggregation strategies are evaluated i.e., MaxP, AvgP, MVoteP, and WSumP.

\begin{figure}[hbt!]
    \centering
    \includegraphics[width=0.6\columnwidth]{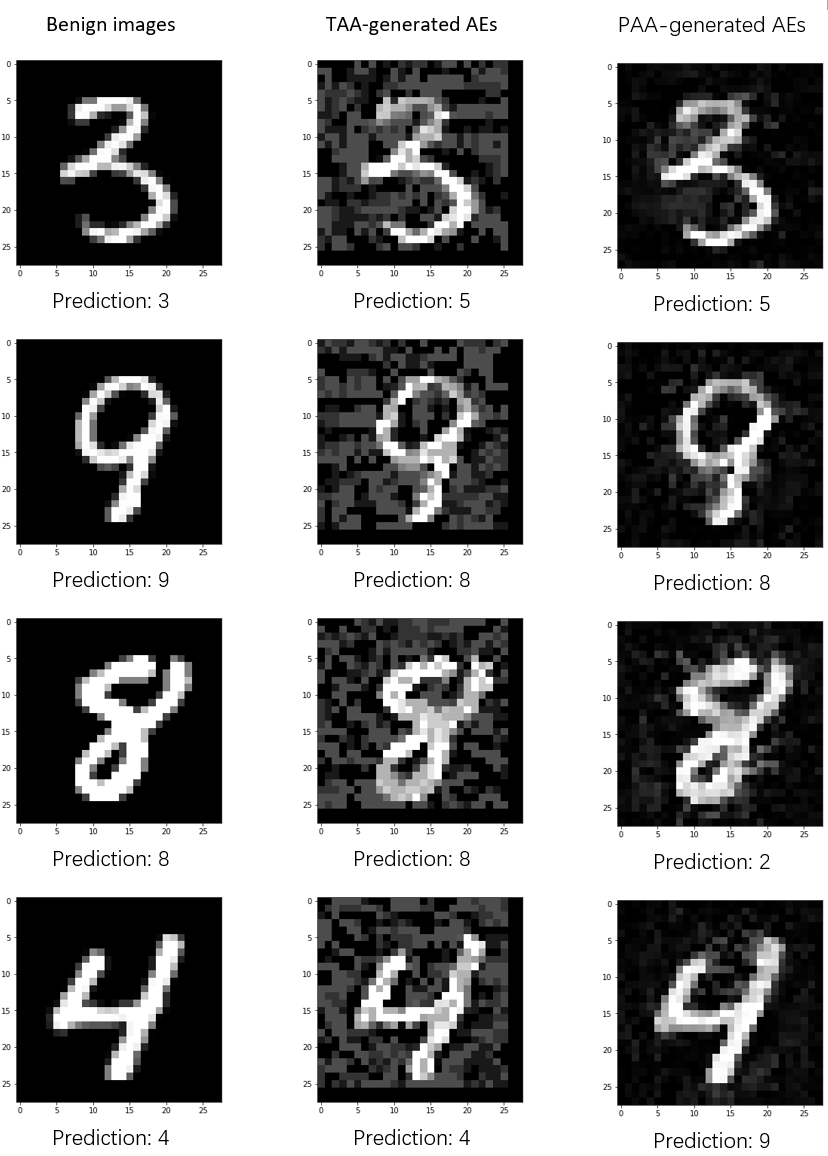}
    \caption{Examples of AEs Generated Using TAA and PAA.}
    \label{fig:ae_comparison}
\end{figure}

We can see from Table~\ref{tab:PAA_results}, for PAA using FGSM, PGD and DeepFool as base attack algorithms, at least one of the four aggregation strategies makes the ensemble accuracy drop to around 20\%.
Under the PGD attack combining WSumP aggregation, the classification accuracy of the transformation-based ensemble drops from 99.70\% on benign samples to 16.65\%.
Some adversarial examples generated using PGD-based TAA and PGD-based PAA are available in Figure~\ref{fig:ae_comparison}.
Unlike TAA, aggregated adversarial perturbations generated by C\&W and DeepFool also have impressive attack effect.
For PAA using C\&W, the ensemble accuracy is lowered to around 70\%.

Also, the perturbation aggregation strategy acts as an important role in our PAA algorithm.
WSumP and AvgP generally outperform the other two aggregation strategies. 
For example, the classification accuracy of ensemble using AVEL ensemble strategy is 95.51\% on FGSM-based PAA AEs using MaxP aggregation strategy, while the classification accuracy is 22.30\% under the same setting except that AvgP aggregation is used.
We observe that, under the same setting, classification accuracy on AEs generated by PAA with AvgP and WSumP as aggregation strategy is similar.
This can be easily explained that for an ensemble with 14 members, 12 of them share the same weight during WSumP according to Equation~\ref{eq:wsp}.
However, this little difference in weights improves the attack effectiveness by dropping the ensemble accuracy 5\% more in some cases when using PGD and DeepFool.
We believe that the attack effectiveness can be further improved by fine-tuning the WSumP weights in Equation~\ref{eq:wsp}, which is though not the main focus in this experiment.

In summary, the TAA and PAA experiments show that the ensemble solely based on reversible transformations cannot provide us with the expected robustness.
The transferability still widely exists among the sub-models trained on data after different reversible transformations, which may be one of the fundamental reasons that such an ensemble is not robust against AEs.
For those attack algorithms whose AEs have less transferability, PAA can assist them to succeed an impressive attack against the reversible ensemble.    

\subsection{Evaluating Ensemble Defence Based on Irreversible Transformations}
To validate the influence of sub-models based on irreversible transformation on the overall robustness of an ensemble, we design an experiment by gradually incorporating more such sub-models into the previously evaluated ensemble of 14 reversible sub-models and measuring their accuracy.
The number of irreversible sub-models $m$ are the multiples of the number 14, i.e., $m \in \{0, 14, 28, 42, 56\}$.
Moreover, we also evaluate the robustness of ensemble of $m$ sub-models solely based on irreversible transformations.

In this experiment, AEs are generated on the original model with TAA attack parameters in Table~\ref{tab:attack_params}.
We generate AEs on the original model for the following two reasons.
First of all, TAA and PAA are not available due to the introduction of irreversible transformation sub-models.
Secondly, it has been demonstrated in the TAA experiment that the AEs generated on the original model can effectively attack the reversible transformation ensemble, which could serve as a baseline. 
We repeat the random drawing and evaluating process for five times for every value of $m$.

The full results on MNIST dataset are shown in Appendix Table~\ref{tab:TAA_athena}.
Regarding C\&W and DeepFool, we have already understood well in Section~\ref{subsec: transferability-results} that their generated AEs have limited transferability among models based on reversible transformations. 
Table~\ref{tab:TAA_athena} further confirms this results by showing that their generated AEs do not affect the ensembles much, no matter the ensemble is based on reversible or irreversible transformations.
However, under FGSM and PGD attacks, we can see that all the ensembles cannot provide expected robustness.
Specifically, we can observe that there is a slight accuracy increase (less than 10\%) when adding the first batch of 14 irreversible sub-models into the ensemble of 14 reversible sub-models.  
However, this increase does not persist as more irreversible sub-models are added. 
Comparing two ensembles of the same number of 14 reversible sub-models and 14 irreversible sub-models, the irreversible ensemble showing slightly advantage in accuracy, i.e., less than 10\% under the FGSM attack and less than 20\% under the PGD attack. 
Also, for the ensemble purly based on irreversible sub-models, increasing the number of sub-models does not increase the ensemble accuracy much.

\begin{figure}[hbt!]
    \centering
    \includegraphics[width=\columnwidth]{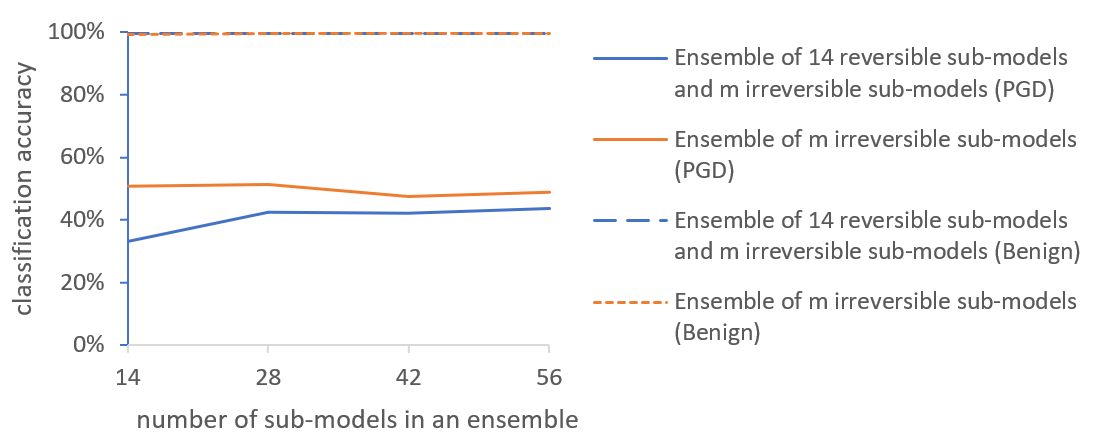}
    \caption{Classification accuracy of ensembles with different number of sub-models predicted with MV strategy.}
    \label{fig:pgd_mv}
\end{figure}

For illustration purpose, we demonstrate the results of ensembles using MV strategy on AEs generated using PGD in Figure~\ref{fig:pgd_mv}. 
The blue line represents the ensemble of 14 sub-models of reversible transformations plus $m$ sub-models of irreversible transformations.
The orange line represents the ensemble of $m$ sub-models solely based on irreversible transformations.
The x-axis is the total number of sub-models in the ensemble.

From Figure~\ref{fig:pgd_mv}, we can observe that classification accuracy under attack increases a little bit as the number of sub-models in the ensemble increases when the ensemble is gradually added in sub-models based on irreversible transformations (i.e., the blue line).
Since the number of sub-models on reversible transformations in this ensemble is fixed, this may imply that sub-models on irreversible transformations bring robustness to the original ensemble with sub-models on reversible transformations only.
For an ensemble of models based on irreversible transformations, the classification accuracy does not change much with respect to the number of sub-models in an ensemble (i.e., the orange line).
The classification accuracy of ensemble solely based on irreversible transformations is always higher than the hybrid ensemble of sub-models on both kinds of transformations, suggesting that sub-models on irreversible transformations may provide more robustness than sub-models on reversible transformations do.
However, the accuracy under attacks is much lower than the benign performance, which indicates that transformation-based ensemble alone may not be able to provide us with enough robustness. 
These observations are also supported by the experimental results on FasionMNIST dataset~\cite{xiao2017fashion} which are available in Appendix Table~\ref{tab:TAA_athena_fmnist}.

\section{Related Work}\label{sec:related-work}
\subsection{Adversarial Machine Learning}\label{subsec:adversarial-machine-learning}
It is well recognised that machine learning algorithms are vulnerable to adversarial examples which are crafted by adding small, human-imperceptible perturbations to legitimate input samples.
There have been various attack algorithms to compute the adversarial perturbations.
For a differentiable machine learning model, attack algorithms can utilise the model gradient information to compute a perturbation which increases the loss function value of the legitimate input to its correct label.
It leads the model to predict a wrong label with a smaller loss function value than that of the correct label~\cite{biggio2013evasion, fgsm, madry2017towards, BIM}.
For non-differentiable classifiers like decision trees and random forests, attackers can apply more complex strategies~\cite{kantchelian2016evasion} or transfer the adversarial example generated with a surrogate model~\cite{russu2016secure}.
Attackers can also use the optimisation-based attack algorithms which are to find the optimised perturbation by maximising or minimising their objective instead of just finding any perturbation that works~\cite{carlini2017towards, chen2017zoo, melis2017deep}.
More intriguingly, there are $L_1$ norm bounded attack algorithms to limit the number of perturbed pixels~\cite{papernot2016limitations, carlini2017towards}, universal adversarial perturbations that work for all examples in the test dataset~\cite{moosavi2017universal,wallace2019universal}, etc.

Recently, the adaptive attack, which is the attack specifically designed for certain defence mechanisms, has been gaining more and more attention due to its impressive attack effect~\cite{carlini2017towards,Carlini}. 
Tramer et al.~\cite{Carlini} evaluated 13 defence methods proposed in the year of 2019 by performing adaptive attacks on them as the evaluation of robustness under worst-case scenario.
Unfortunately, all of these 13 defence methods fail to protect victim model from corresponding adaptive attacks while majority of adaptive attacks on them are iterative attack algorithms like PGD with multiple random starts and larger number of iterations.
Our paper borrows the core idea of the classical attack algorithms and proposes adaptive attack algorithms targeting the transformation-based ensemble defence, fully considering the learnt lessons about the importance of evaluating under sufficient strong settings~\cite{Carlini}.

\subsection{Ensemble Defence}\label{subsec:ensemble-defenses}
There is a continuous arms race between the machine learning attack and defence. 
It is of great significance to defend machine learning models against such attacks so that the models can produce trustworthy output that we can rely on.  
Researchers have put a large amount of effort in defending machine learning models\cite{ECNN}\cite{adp}\cite{papernot2016distillation}, among which the ensemble defence is an important branch.

Ensemble as a defence method is an easy but efficient way to protect models from adversarial attacks especially under black-box scenarios when the ensemble details are unknown to attackers.
There are ensembles promoting the diversity among sub-models~\cite{adp}, using different precision of weights and activation functions in sub-models~\cite{sen2020empir}, training an ensemble of binary-classifiers with sufficient diversity and redundancy~\cite{ECOC}, combining sub-models trained on data after different transformations~\cite{meng2020ensembles}, etc.
It is shown that even simple combination of undefended classifiers with different random noise has similar robustness to the adversarial training~\cite{rse, carlini2017towards} and advanced ensembles with guaranteed diversity~\cite{adp, ECOC, ECNN, meng2020ensembles} demonstrate superior performance.
However, in the attack and defence arms race, some defences are soon broken by adaptive attacks~\cite{adp,he2017adversarial, sen2020empir, Carlini}.
In order to better understand the effectiveness of transformation-based ensemble defence~\cite{meng2020ensembles}, we utilise transferability analysis and propose two adaptive attack algorithms, i.e., TAA and PAA. 
The results reveal that the transformation-based ensemble defence can provide some but not enough robustness against AEs.
The gained robustness is mainly from the irreversible transformation sub-models, and increasing the number ensemble sub-models does not increase the ensemble robustness.

\section{Conclusion}\label{sec:conclusion}
In this work, we designed adaptive attacks to evaluate the robustness of the transformation-based ensemble defence and conducted controlled experiments to analyse the reason for such robustness by considering transformation types. 
The experimental results indicated that transformation-based ensemble defence cannot provide expected robustness.
Though our experimental findings are not positive in trusting machine learning models with critical tasks, they provide us with more detailed understanding why we cannot rely on such defence.
There is a necessity of continuous defence effort in this machine learning attack and defence arms race.

\bibliographystyle{ACM-Reference-Format}
\bibliography{ensemble2020}
\newpage
    \appendix
\section{Appendix}
    
\begin{table}[hb]
    \centering
    \caption{Classification Accuracy of Ensembles on MNIST with Different Sub-Models Under Adversarial Attacks. ``r'' refers to reversible transformation, ``i'' refers to irreversible transformation, and the number in front of them indicates the number of such sub-models in the ensemble.}
    \label{tab:TAA_athena}
    \resizebox{\columnwidth}{!}{%
    \begin{tabular}{|l|l|l|l|l|l|l|}
    \hline
     &  & \textbf{RD} & \textbf{MV} & \textbf{T2MV} & \textbf{AVEP} & \textbf{AVEL} \\ \hline
     \multirow{9}{*}{\textbf{Benign}} & \textbf{14r} & 99.36\% & 99.70\% & 96.48\% & 99.68\% & 99.68\% \\ \cline{2-7} 
     & \textbf{14r+14i} & 99.27\% & 99.70\% & 98.46\% & 99.70\% & 99.70\% \\ \cline{2-7} 
     & \textbf{14r+28i} & 99.10\% & 99.70\% & 99.03\% & 99.70\% & 99.71\% \\ \cline{2-7} 
     & \textbf{14r+42i} & 99.38\% & 99.70\% & 99.16\% & 99.70\% & 99.70\% \\ \cline{2-7} 
     & \textbf{14r+56i} & 99.38\% & 99.70\% & 99.30\% & 99.69\% & 99.71\% \\ \cline{2-7} 
     & \textbf{14i} & 99.00\% & 99.66\% & 97.16\% & 99.68\% & 99.69\% \\ \cline{2-7} 
 & \textbf{28i} & 99.18\% & 99.70\% & 98.25\% & 99.68\% & 99.69\% \\ \cline{2-7} 
 & \textbf{42i} & 99.23\% & 99.69\% & 98.89\% & 99.68\% & 99.68\% \\ \cline{2-7} 
 & \textbf{56i} & 99.17\% & 99.70\% & 99.06\% & 99.71\% & 99.70\% \\ \hline
 \multirow{9}{*}{\textbf{FGSM}} & \textbf{14r} & 58.68\% & 69.44\% & 66.84\% & 69.24\% & 68.52\% \\ \cline{2-7} 
 & \textbf{14r+14i} & 66.56\% & 73.59\% & 70.43\% & 72.85\% & 72.17\% \\ \cline{2-7} 
         & \textbf{14r+28i} & 64.82\% & 75.56\% & 72.60\% & 74.94\% & 73.69\% \\ \cline{2-7} 
         & \textbf{14r+42i} & 55.64\% & 75.26\% & 72.20\% & 74.32\% & 73.23\% \\ \cline{2-7} 
         & \textbf{14r+56i} & 57.26\% & 75.99\% & 72.94\% & 75.14\% & 73.71\% \\ \cline{2-7} 
         & \textbf{14i} & 66.64\% & 76.08\% & 72.18\% & 74.70\% & 73.31\% \\ \cline{2-7} 
         & \textbf{28i} & 67.78\% & 76.28\% & 74.09\% & 75.68\% & 74.42\% \\ \cline{2-7} 
         & \textbf{42i} & 61.28\% & 75.84\% & 72.00\% & 74.50\% & 72.29\% \\ \cline{2-7} 
         & \textbf{56i} & 59.71\% & 77.77\% & 75.78\% & 77.31\% & 75.56\% \\ \hline
        \multirow{9}{*}{\textbf{PGD}} & \textbf{14r} & 34.26\% & 33.24\% & 34.84\% & 31.54\% & 30.62\% \\ \cline{2-7} 
         & \textbf{14r+14i} & 55.30\% & 42.36\% & 41.49\% & 41.71\% & 41.76\% \\ \cline{2-7} 
         & \textbf{14r+28i} & 39.51\% & 41.98\% & 41.26\% & 41.61\% & 42.59\% \\ \cline{2-7} 
         & \textbf{14r+42i} & 43.25\% & 43.57\% & 41.50\% & 43.31\% & 42.27\% \\ \cline{2-7} 
         & \textbf{14r+56i} & 33.42\% & 45.08\% & 41.80\% & 44.57\% & 43.58\% \\ \cline{2-7} 
         & \textbf{14i} & 36.68\% & 50.81\% & 46.44\% & 49.74\% & 46.84\% \\ \cline{2-7} 
         & \textbf{28i} & 42.56\% & 51.22\% & 46.64\% & 50.63\% & 48.44\% \\ \cline{2-7} 
         & \textbf{42i} & 37.14\% & 47.54\% & 43.22\% & 46.79\% & 44.84\% \\ \cline{2-7} 
         & \textbf{56i} & 43.60\% & 48.92\% & 43.62\% & 48.04\% & 45.15\% \\ \hline
        \multirow{9}{*}{\textbf{C\&W}} & \textbf{14r} & 97.00\% & 98.58\% & 92.82\% & 98.86\% & 98.82\% \\ \cline{2-7} 
         & \textbf{14r+14i} & \multicolumn{1}{r|}{96.87\%} & \multicolumn{1}{r|}{99.18\%} & \multicolumn{1}{r|}{96.65\%} & \multicolumn{1}{r|}{99.26\%} & \multicolumn{1}{r|}{99.28\%} \\ \cline{2-7} 
         & \textbf{14r+28i} & \multicolumn{1}{r|}{97.20\%} & \multicolumn{1}{r|}{99.32\%} & \multicolumn{1}{r|}{97.38\%} & \multicolumn{1}{r|}{99.36\%} & \multicolumn{1}{r|}{99.40\%} \\ \cline{2-7} 
         & \textbf{14r+42i} & \multicolumn{1}{r|}{95.83\%} & \multicolumn{1}{r|}{99.37\%} & \multicolumn{1}{r|}{98.05\%} & \multicolumn{1}{r|}{99.38\%} & \multicolumn{1}{r|}{99.40\%} \\ \cline{2-7} 
         & \textbf{14r+56i} & \multicolumn{1}{r|}{97.97\%} & \multicolumn{1}{r|}{99.66\%} & \multicolumn{1}{r|}{97.76\%} & \multicolumn{1}{r|}{99.66\%} & \multicolumn{1}{r|}{99.66\%} \\ \cline{2-7} 
         & \textbf{14i} & 97.78\% & 99.59\% & 94.68\% & 99.63\% & 99.58\% \\ \cline{2-7} 
         & \textbf{28i} & 82.78\% & 99.63\% & 96.15\% & 99.64\% & 99.65\% \\ \cline{2-7} 
         & \textbf{42i} & 96.92\% & 99.63\% & 97.41\% & 99.66\% & 99.65\% \\ \cline{2-7} 
         & \textbf{56i} & 98.01\% & 99.65\% & 97.94\% & 99.66\% & 99.65\% \\ \hline
        \multirow{9}{*}{\textbf{DeepFool}} & \textbf{14r} & 97.90\% & 99.50\% & 92.30\% & 99.52\% & 99.62\% \\ \cline{2-7} 
         & \textbf{14r+14i} & 97.94\% & 99.66\% & 96.19\% & 99.67\% & 99.66\% \\ \cline{2-7} 
         & \textbf{14r+28i} & 98.36\% & 99.64\% & 97.40\% & 99.67\% & 99.67\% \\ \cline{2-7} 
         & \textbf{14r+42i} & 98.21\% & 99.65\% & 97.66\% & 99.66\% & 99.65\% \\ \cline{2-7} 
         & \textbf{14r+56i} & 96.37\% & 99.33\% & 98.17\% & 99.36\% & 99.39\% \\ \cline{2-7} 
         & \textbf{14i} & 97.18\% & 99.12\% & 93.68\% & 99.22\% & 99.22\% \\ \cline{2-7} 
     & \textbf{28i} & 96.37\% & 99.32\% & 96.84\% & 99.36\% & 99.40\% \\ \cline{2-7} 
     & \textbf{42i} & 96.96\% & 99.34\% & 97.48\% & 99.38\% & 99.37\% \\ \cline{2-7} 
     & \textbf{56i} & 96.26\% & 99.37\% & 97.60\% & 99.38\% & 99.36\% \\ \hline
    \end{tabular}%
    }
\end{table}

\begin{table}[hb]
\centering
\caption{Classification Accuracy of Ensembles on FashionMNIST with Different Sub-Models Under Adversarial Attacks. “r” refers to reversible transformation, “i” refers to irreversible transformation, and the number in front of them indicates the number of such sub-models in the ensemble.}
\label{tab:TAA_athena_fmnist}
\resizebox{\columnwidth}{!}{%
\begin{tabular}{|l|l|l|l|l|l|l|}
\hline
\multicolumn{2}{|l|}{} & \textbf{RD} & \textbf{MV} & \textbf{T2MV} & \textbf{AVEP} & \textbf{AVEL} \\ \hline
\multirow{9}{*}{\textbf{Benign}} & \textbf{14r} & 90.66\% & 93.16\% & 86.90\% & 93.22\% & 93.56\% \\ \cline{2-7} 
 & \textbf{14r+14i} & 89.99\% & 93.13\% & 88.78\% & 93.20\% & 93.22\% \\ \cline{2-7} 
 & \textbf{14r+28i} & 90.24\% & 92.81\% & 89.22\% & 92.89\% & 92.74\% \\ \cline{2-7} 
 & \textbf{14r+42i} & 88.41\% & 92.82\% & 89.45\% & 92.88\% & 92.77\% \\ \cline{2-7} 
 & \textbf{14r+56i} & 89.23\% & 92.75\% & 89.60\% & 92.81\% & 92.79\% \\ \cline{2-7} 
 & \textbf{14i} & 90.21\% & 92.08\% & 85.06\% & 92.27\% & 92.09\% \\ \cline{2-7} 
 & \textbf{28i} & 87.75\% & 92.26\% & 87.04\% & 92.37\% & 92.22\% \\ \cline{2-7} 
 & \textbf{42i} & 89.40\% & 92.28\% & 88.33\% & 92.40\% & 92.34\% \\ \cline{2-7} 
 & \textbf{56i} & 90.32\% & 92.56\% & 88.56\% & 92.59\% & 92.54\% \\ \hline
\multirow{9}{*}{\textbf{FGSM}} & \textbf{14r} & 22.34\% & 24.42\% & 23.48\% & 24.44\% & 25.16\% \\ \cline{2-7} 
 & \textbf{14r+14i} & 20.43\% & 24.00\% & 22.64\% & 24.36\% & 24.74\% \\ \cline{2-7} 
 & \textbf{14r+28i} & 20.72\% & 25.49\% & 23.67\% & 25.60\% & 26.61\% \\ \cline{2-7} 
 & \textbf{14r+42i} & 23.02\% & 25.20\% & 23.33\% & 25.45\% & 26.56\% \\ \cline{2-7} 
 & \textbf{14r+56i} & 20.43\% & 25.54\% & 23.87\% & 25.85\% & 27.11\% \\ \cline{2-7} 
 & \textbf{14i} & 22.94\% & 23.55\% & 21.93\% & 23.85\% & 23.87\% \\ \cline{2-7} 
 & \textbf{28i} & 20.26\% & 25.49\% & 23.12\% & 25.87\% & 25.80\% \\ \cline{2-7} 
 & \textbf{42i} & 20.04\% & 24.82\% & 23.00\% & 25.20\% & 25.86\% \\ \cline{2-7} 
 & \textbf{56i} & 20.03\% & 24.19\% & 22.21\% & 24.33\% & 25.31\% \\ \hline
\multirow{9}{*}{\textbf{PGD}} & \textbf{14r} & 9.96\% & 9.20\% & 8.84\% & 9.62\% & 9.32\% \\ \cline{2-7} 
 & \textbf{14r+14i} & 15.12\% & 10.08\% & 9.18\% & 10.39\% & 11.73\% \\ \cline{2-7} 
 & \textbf{14r+28i} & 20.06\% & 10.72\% & 9.63\% & 11.12\% & 12.41\% \\ \cline{2-7} 
 & \textbf{14r+42i} & 14.03\% & 10.36\% & 9.30\% & 10.76\% & 12.28\% \\ \cline{2-7} 
 & \textbf{14r+56i} & 10.33\% & 10.68\% & 9.44\% & 11.00\% & 13.42\% \\ \cline{2-7} 
 & \textbf{14i} & 11.24\% & 11.94\% & 10.63\% & 12.61\% & 13.64\% \\ \cline{2-7} 
 & \textbf{28i} & 17.78\% & 11.51\% & 9.91\% & 11.94\% & 13.12\% \\ \cline{2-7} 
 & \textbf{42i} & 13.04\% & 12.63\% & 10.66\% & 12.92\% & 13.86\% \\ \cline{2-7} 
 & \textbf{56i} & 12.21\% & 10.77\% & 9.55\% & 11.05\% & 12.24\% \\ \hline
\multirow{9}{*}{\textbf{C\&W}} & \textbf{14r} & 90.16\% & 92.82\% & 83.84\% & 93.02\% & 93.14\% \\ \cline{2-7} 
 & \textbf{14r+14i} & 85.72\% & 92.80\% & 86.41\% & 92.98\% & 92.80\% \\ \cline{2-7} 
 & \textbf{14r+28i} & 86.43\% & 92.80\% & 87.77\% & 92.95\% & 92.70\% \\ \cline{2-7} 
 & \textbf{14r+42i} & 87.22\% & 92.71\% & 88.08\% & 92.80\% & 92.66\% \\ \cline{2-7} 
 & \textbf{14r+56i} & 87.65\% & 92.51\% & 88.02\% & 92.61\% & 92.54\% \\ \cline{2-7} 
 & \textbf{14i} & 86.51\% & 91.55\% & 82.64\% & 91.74\% & 91.43\% \\ \cline{2-7} 
 & \textbf{28i} & 85.48\% & 91.96\% & 85.67\% & 92.04\% & 91.87\% \\ \cline{2-7} 
 & \textbf{42i} & 86.47\% & 92.17\% & 86.55\% & 92.24\% & 91.99\% \\ \cline{2-7} 
 & \textbf{56i} & 84.65\% & 92.01\% & 87.20\% & 92.15\% & 92.10\% \\ \hline
\multirow{9}{*}{\textbf{DeepFool}} & \textbf{14r} & 17.94\% & 18.60\% & 16.68\% & 18.63\% & 18.62\% \\ \cline{2-7} 
 & \textbf{14r+14i} & 88.66\% & 92.97\% & 85.52\% & 93.02\% & 93.05\% \\ \cline{2-7} 
 & \textbf{14r+28i} & 88.40\% & 92.59\% & 85.41\% & 92.81\% & 92.69\% \\ \cline{2-7} 
 & \textbf{14r+42i} & 87.92\% & 92.70\% & 86.42\% & 92.76\% & 92.68\% \\ \cline{2-7} 
 & \textbf{14r+56i} & 88.20\% & 92.36\% & 86.20\% & 92.46\% & 92.40\% \\ \cline{2-7} 
 & \textbf{14i} & 87.36\% & 91.49\% & 80.62\% & 91.68\% & 91.26\% \\ \cline{2-7} 
 & \textbf{28i} & 72.94\% & 92.01\% & 82.40\% & 92.12\% & 92.00\% \\ \cline{2-7} 
 & \textbf{42i} & 88.21\% & 92.14\% & 85.92\% & 92.29\% & 92.22\% \\ \cline{2-7} 
 & \textbf{56i} & 87.34\% & 91.76\% & 84.03\% & 91.88\% & 91.96\% \\ \hline
\end{tabular}%
}
\end{table}

\end{document}